\documentclass[runningheads]{llncs}
\usepackage{graphicx}
\usepackage{appendix}
\usepackage{amssymb}
\usepackage{amsbsy}
\usepackage{amsmath}
\usepackage{xcolor}
\usepackage{listings}
\usepackage{mdframed}
\usepackage{stmaryrd}
\usepackage{url}

\usepackage[normalem]{ulem}

\lstdefinelanguage{Isabelle}
{
	basicstyle=\ttfamily\small,
	keywords=[1]{
		definition, lemma, using, by, unfolding, named_theorems, declare, consts, axiomatization, typedecl, type_synonym, oops, theorem, abbreviation, begin, theory, imports, end, sledgehammer
	},
	keywordstyle=[1]\bfseries,
	keywords=[2]{where, infixr, assumes, shows, and},
	keywordstyle=[2]\bfseries,
	keywords=[3]{v, p, r},
	keywordstyle=[3],
	keywords=[4]{add},
	keywordstyle=[4],
	sensitive=false,
	morestring=[b]',
	morecomment=[l]{--}
}

\newcommand{\RNum}[1]{\uppercase\expandafter{\romannumeral #1\relax}}

\newcommand{\hl}[1]{#1}

\begin{document}
\title{Public  Announcement Logic in HOL}
\author{Sebastian Reiche\hl{\orcidID{0000-0002-2555-999X}} \and
Christoph Benzm\"uller\hl{\orcidID{0000-0002-3392-3093}}}
%\author{Sebastian Reiche \and Christoph Benzm\"uller}
%
\authorrunning{S. Reiche and C. Benzm\"uller}
\institute{Freie Universit\"at Berlin, Berlin, Germany\\
\email{\{sebastian.reiche,c.benzmueller\}@fu-berlin.de}}
\maketitle              % typeset the header of the contribution
\begin{abstract}
A shallow semantical embedding for public announcement logic with relativized common knowledge is presented. 
This embedding enables the first-time automation of this logic with off-the-shelf theorem provers for classical higher-order logic. It is demonstrated (i) how meta-theoretical studies can be automated this way, and (ii) how non-trivial reasoning in the target logic (public announcement logic), required e.g.~to obtain a convincing encoding and automation of the wise men puzzle, can be realized.
Key to the presented semantical embedding---in contrast, e.g.,~to related work on the  semantical embedding of normal modal logics---is that evaluation domains are modeled explicitly and treated as additional parameter in the encodings of the constituents of the embedded target logic, while they were  previously implicitly shared between meta logic and target logic.

\end{abstract}
\section{Introduction}
%\hl{Expand the abstract from above.  Refer also to \cite{J41} and \cite{J44} for automation of the wise men puzzle and say what we are here doing better. Mention also all other attempts that are known to automate the wise men puzzle on the computer. Mention also other attempts to automate the meta-theory of public announcement logic on the computer (if any).}

Previous work has studied the application of a 
universal (meta-)logical reasoning approach \cite{J41,J44} 
for solving  a prominent riddle in epistemic reasoning, known as the \textit{wise men puzzle}, on the computer \cite{J44}. The solution presented there puts a particular emphasis on the adequate modeling of (ordinary) common knowledge and it also illustrates the elegance and the practical relevance of the shallow semantical embedding approach (in classical higher-order logic) \cite{J41}, when being utilized within modern proof assistant systems such as Isabelle/HOL \cite{isabelle}. However, this work nevertheless falls short, since it did not convincingly  address the interaction dynamics between the involved agents.  To do so, we extend and adapt in this student paper the universal (meta-)logical reasoning approach for public announcement logic and we demonstrate how it can be adapted to achieve a convincing encoding and automation of the wise men puzzle in Isabelle/HOL that also captures the interaction dynamics of the wise men puzzle scenario.
In more general terms, we present the first automation of 
public announcement logic with relativized common knowledge, and we demonstrate that, and how, this logic can be seen and elegantly handled as a fragment of classical higher-order logic. 
Key to the presented extension of the shallow semantical embedding approach is that the evaluation domains of the embedded 
target logic (public announcement logic with relativized common knowledge) are no longer implicitly shared with the meta-logic (classical higher-order logic), but they are now
explicitly modeled as an additional parameter in the encoding of the embedded logics constituents. 

% The sources of our contribution are available in subfolder \texttt{Public-Announcement-Logic} at \url{www.logikey.org}.

This paper is structured as follows:
\textsection 2 briefly recaps classical higher-order logic (Church's type theory), and
\textsection 3 sketches public announcement logic with relativized common knowledge. 
The main contributions of this paper are then presented in \textsection 4, where a shallow semantical embedding of public announcement logic in classical higher-order logic is studied.
In \textsection 5 the newly acquired embedding is tested and applied to achieve an encoding and automation of the prominent wise men puzzle.
%the introduced framework.
\hl{\textsection 6 discusses related work and \textsection 7 concludes the paper.}

\section{Classical Higher-Order Logic}
We briefly recap classical higher-order logic (HOL), respectively Church's  \textit{simple theory of types} \cite{church1940,J43}, which is a logic defined on top of the simply typed lambda calculus.
The presentation is partly adapted from  Benzm\"uller~\cite{J31}. For further information on the syntax and semantics of HOL we refer to \cite{J6}.

\paragraph{Syntax of HOL.}
%\paragraph{Types.} 
We start out with defining 
%To formulate higher-order logic, we first need a set of types $\mathcal{T}$.
the set $\mathcal{T}$ of \textit{simple types} by the following abstract grammar:
	$\alpha , \beta := o \ | \ i \ | \ (\alpha \rightarrow \beta)$.
	Type $o$ denotes a bivalent set of truth values, containing \textit{truth} and \textit{falsehood}, and $i$ denotes a non-empty set of individuals. Further base types are optional.
	$\rightarrow$ is the function type constructor, such that $(\alpha \rightarrow \beta) \in \mathcal{T}$ whenever $\alpha, \beta \in \mathcal{T}$. We may generally omit parentheses.
	
%\paragraph{Terms.} 
The \textit{terms} of HOL are defined  by the following abstract grammar:
$$s,t := p_\alpha \ | \ X_\alpha \ | \ (\lambda x_\alpha . s_\beta)_{\alpha \rightarrow \beta} \ | \ (s_{\alpha\rightarrow\beta}t_\alpha)_\beta$$
where $\alpha, \beta, o \in \mathcal{T}$. The $p_\alpha \in C_\alpha$ are typed constants  and the $X_\alpha \in V_\alpha$ are typed variables (distinct from the $p_\alpha$).
	%Complex typed HOL terms are thus constructed via $\lambda$-abstraction and function application.
	If $s_{\alpha \rightarrow \beta}$ and $t_\alpha$ are HOL terms of types $\alpha \rightarrow \beta$ and $\alpha$, respectively, then $(s_{\alpha \rightarrow \beta}t_\alpha)_\beta$, called \textit{application}, is an HOL term of type $\beta$.
	If $X_\alpha \in V_\alpha$ is a typed variable symbol and $s_\beta$ is an HOL term of type $\beta$, then $(\lambda X_\alpha s_\beta)_{\alpha \rightarrow \beta}$, called \textit{abstraction}, is an HOL term of type $\alpha \rightarrow \beta$.
	The type of each term is given as a subscript.
	We call terms of type $o$ \textit{formulas}.
	As \textit{primitive logical connectives} we choose $\neg_{o \rightarrow o}, \vee_{o\rightarrow o\rightarrow o}$, $=_{\alpha \rightarrow \alpha \rightarrow \alpha }$ and $\Pi_{(\alpha \rightarrow o) \rightarrow o}$.
	Other logical connectives can be introduced as abbreviations\hl{; e.g. $\longrightarrow_{o\rightarrow o\rightarrow o} = \lambda y_o. \lambda y_o. \neg x \vee y$.}

% A variable can either occur bound by a $\lambda$ or free. $free(s)$ is used to denote the set of variables with free occurrences in $s$.

%\textit{Substitution}
%	The substitution of a term $s_\alpha$ for a variable $X_\alpha$ in a term $t_\beta$ is denoted as $[s/X]t$.We consider $\alpha$-conversion implicitly, making substitutions capture-avoiding.
	% $\beta$ and $\eta$-reduction are defined as usual: $(\beta)$ \hspace*{.4em}$(\lambda X.s)t = [t/X]s$ $(\eta)$ $(\lambda X.s X) = s$\ \ ($X \not \in free(s)$)

\paragraph{Semantics of HOL.}
%We now define the necessary structure and conditions for satisfiabilty in HOL, in such ways that give meaning to logical constructs.
	A \textit{frame} $\mathcal{D}$ for HOL is a collection $\{\mathcal{D}_\alpha\}_{\alpha \in T}$ of nonempty sets $\mathcal{D}_\alpha$, such that $\mathcal{D}_o = \{T, F\}$ (for true and false).
	$\mathcal{D}_i$ is chosen freely and $\mathcal{D}_{\alpha \rightarrow \beta}$ are collections of functions mapping $\mathcal{D}_\alpha$ into $\mathcal{D}_\beta$.

	A \textit{model} for HOL is a tuple $\mathcal{M} = \langle \mathcal{D}, I \rangle$, where $\mathcal{D}$ is a frame, and $I$ is a family of typed interpretation functions mapping constant symbols $p_\alpha \in C_\alpha$ to appropriate elements of $\mathcal{D}_\alpha$, called the \textit{denotation} of $p_\alpha$.
	The logical connectives $\neg, \vee, \Pi$ and $=$ are always given their expected standard denotations:
	\[\begin{tabular}{lll}
		$I(\neg_{o\rightarrow o})$ & = \textit{not} $\in \mathcal{D}_{o \rightarrow o}$ &s.t. \textit{not}(T) = F and \textit{not}(F) = T\\
		$I(\vee_{o \rightarrow o \rightarrow o})$ & = \textit{or} $\in \mathcal{D}_{o \rightarrow o \rightarrow o}$ &s.t. \textit{or}(a,b) = T iff (a = T or b = T) \\
		$I(=_{\alpha \rightarrow \alpha \rightarrow o})$ & = \textit{id} $\in \mathcal{D}_{\alpha \rightarrow \alpha \rightarrow o}$ &s.t. for all a,b $\in \mathcal{D}_\alpha$, \textit{id}(a,b) = T\\\
		&&\quad iff a is identical to b.\\
		$I(\Pi_{(\alpha \rightarrow o) \rightarrow o})$ & = \textit{all} $\in \mathcal{D}_{(\alpha \rightarrow o) \rightarrow o}$ & s.t. for all $s \in \mathcal{D}_{\alpha \rightarrow o}$, \textit{all}(s) = T\\
		&&\quad iff s(a) = T for all a $\in \mathcal{D}_\alpha$ 
	\end{tabular}\]
	A \textit{variable assignment g} maps variables $X_\alpha$ to elements in $\mathcal{D}_\alpha$.
	$g[d/W]$ denotes the assignment that is identical to $g$, except for variable $W$, which is now mapped to $d$.
	
	The \textit{denotation} \hl{$\llbracket s_\alpha\rrbracket^{M,g}$} of an HOL term $s_\alpha$ on a model $\mathcal{M} = \langle \mathcal{D}, I \rangle$ under assignment $g$ is an element $d \in \mathcal{D}_\alpha$ defined in the following way:
	\[
	\begin{tabular}{lll}
		$\llbracket p_\alpha\rrbracket^{\mathcal{M}, g}$ & = & $I(p_\alpha)$\\
		$\llbracket X_\alpha\rrbracket^{\mathcal{M}, g}$ & = & $g(X_\alpha)$\\
		$\llbracket (s_{\alpha \rightarrow \beta}t_\alpha)_\beta\rrbracket^{\mathcal{M}, g}$ & = & $\llbracket s_{\alpha \rightarrow \beta}\rrbracket^{\mathcal{M}, g}(\llbracket t_\alpha\rrbracket^{\mathcal{M}, g})$\\
		$\llbracket (\lambda X_\alpha s_\beta)_{\alpha \rightarrow \beta}\rrbracket^{\mathcal{M}, g}$ & = & the function $f$ from $\mathcal{D}_\alpha$ to $\mathcal{D}_\beta$\\
		&&\quad s.t. $f(d) = \llbracket s_\beta \rrbracket^{\mathcal{M},g[d/X_\alpha]}$ for all $d \in \mathcal{D}_\alpha$
	\end{tabular}
    \]
	In a \textit{standard model} a domain $\mathcal{D}_{\alpha \rightarrow \beta}$ is defined as the set of all total functions from $\mathcal{D}_\alpha$ to $\mathcal{D}_\beta$:
%	All domains for the function types are defined inductively: 
    $\mathcal{D}_{\alpha \rightarrow \beta} = \{ f \ |\ f : \mathcal{D}_\alpha \rightarrow \mathcal{D}_\beta \}$. 
	In a \textit{Henkin model} (or general model) \cite{henkin1950} function spaces are not necessarily required to be the full set of functions: $\mathcal{D}_{\alpha \rightarrow \beta} \subseteq \{ f \ |\ f : \mathcal{D}_\alpha \rightarrow \mathcal{D}_\beta \}$.  However, we require that the valuation function remains total, so that every term denotes \sout{\hl{(Denotatpflicht)}}.

%\paragraph{Validity.}
	A HOL formula $s_o$ is \textit{valid in an Henkin model $\mathcal{M}$ under assignment $g$} if and only if \hl{$\llbracket s_o\rrbracket^{\mathcal{M},g}$}{$=T$}; also denoted by $\mathcal{M},g \models^\texttt{HOL} s_o$.
	An HOL formula $s_o$ is called \textit{valid in $\mathcal{M}$}, denoted by $\mathcal{M} \models^\texttt{HOL} s_o$, iff $\mathcal{M},g \models^\texttt{HOL}s_o$ for all assignments $g$.
	Moreover, a formula $s_o$ is called \textit{valid}, denoted by $\models^\texttt{HOL}s_o$, if and only if $s_o$ is valid in all Henkin models $\mathcal{M}$.

Due to G\"odel \cite{goedel1931} a sound and complete mechanization of HOL with standard semantics cannot be achieved. For HOL with Henkin semantics sound and complete calculi exist; cf.~e.g.~\cite{B5,J6} and the references therein.

Each standard model is obviously also a Henkin model.
Consequently, when a HOL formula is Henkin-valid, it is also valid in all standard models.

%All that is required is that $\mathcal{D}_{\alpha \rightarrow \beta}$ has enough members that any well-formed formula can be evaluated. We have $\mathcal{D}_{\alpha \rightarrow \beta} \subseteq \{ f \ |\ f : \mathcal{D}_\alpha \rightarrow \mathcal{D}_\beta \}$ such that the valuation function denotes for every term.

%Henkin semantics have been proven sound and complete in \cite{andrews1972}.
%Thus, Henkin semantics (theoretically) allows a complete mechanization.
%Hence, if a proof for a formula $s$ can be found with a Henkin-sound theorem prover for HOL, then we know that $s$ is also a theorem in standard semantics.
%For this reason, Henkin semantics is assumed in the remainder.

\section{Public Announcement Logic}

%TODO: Kripke / epistemic model einheitlich

The most important concepts and definitions of a public announcement logic (PAL) with relativized common knowledge are depicted.
For more details we refer to \cite{doi:10.1111/phc3.12059,van2015introduction}.

Before exploring these definitions some general descriptions of the modeling approach might be worthwhile.
We use a graph-theoretical structure, called \textit{epistemic models}, to represent knowledge.
Epistemic models describe situations in terms of possible worlds.
A world represents one possibility about how the current situation can be.
Each agent is assumed to entertain a number of these possibilities.
Knowledge is described using an accessibility relation between worlds, rather than directly representing the \hl{agent's} information.

Let $\mathcal{A}$ be a set of agents and $\mathcal{P}$ a set of atomic propositions.
Atomic \hl{propositions} are intended to describe ground facts.
We use a set $W$ to denote possible worlds and a valuation function $V:\mathcal{P}\rightarrow \wp (W)$ that assigns a set of worlds to each atomic proposition. Vice versa, we may identify each world with the set of propositions that are validated in them.

\begin{definition}[\textbf{Epistemic Model}]
	Let $\mathcal{A}$ be a (finite) set of agents and $\mathcal{P}$ a (finite or countable) set of atomic propositions.
	An \textit{Epistemic Model} is a triple	$\mathcal{M} = \langle W, \{ R_i \}_{i \in \mathcal{A}}, V \rangle$ where $W \not = \emptyset,\ R_i \subseteq W \times W$ is an accessibility relation (for each $i \in \mathcal{A}$), and $V: \mathcal{P} \rightarrow \wp(W)$ is a valuation function ($\wp(W)$ is the powerset of $W$).
\end{definition}

\noindent
Information of agent $i$ at world $w$ can now be defined as: $R_i(w) = \{v \in W\ | \ wR_iv \}$.
Having a separate (accessibility) relation for each agent enables them to have their own viewpoints.

Next, we introduce the syntax of our base epistemic logic as the set of sentences generated by the following grammar (where $p \in \mathcal{P}$ and $i \in \mathcal{A}$):
\[ \varphi, \psi := p \ | \ \neg \varphi \ | \ \varphi \vee \psi \ | \ K_i \varphi \]
%The interpretation of $\mathcal{L}_\texttt{K}$ uses the following truth definition:

\noindent
\hl{We also introduce the abbreviations $\varphi \wedge \psi := \neg(\neg \varphi \vee \neg \psi)$ and $\varphi \rightarrow \psi := \neg \varphi \vee \psi$.}

\begin{definition}[\textbf{Truth at world $w$}] \sloppy
	Given an epistemic model $\mathcal{M} = \langle W, \{ R_i \}_{i \in \mathcal{A}}, V \rangle$.
	For each $w \in W, \varphi$ \emph{is true at world $w$}, denoted $\mathcal{M}, w \models \varphi$, is defined inductively as follows:

\vspace{.35em}
\begin{tabular}{lll}
	$\mathcal{M},w \models p$ & iff & $w \in V(p)$\\
	$\mathcal{M},w \models \neg \varphi$ & iff & $\mathcal{M}, w \not \models \varphi$\\
	$\mathcal{M},w \models \varphi \vee \psi$ & iff & $\mathcal{M},w \models \varphi$ or $\mathcal{M},w \models \psi$\\
	$\mathcal{M},w \models K_i \varphi$ & iff & for all $v \in W$, if $wR_iv$ then $\mathcal{M},v \models \varphi$
\end{tabular}
\end{definition}

\noindent	
The formula $K_i\varphi$  expresses that "Agent $i$ knows $\varphi$".
This describes knowledge as an all-or-nothing definition.
If we postulate that agent $i$ knows $\varphi$, we say that $\varphi$ is true throughout all worlds in agents $i$'s range of considerations.

\textit{Satisfiabilty} of a formula $\varphi$ for a model $\mathcal{M} = \langle W, \{ R_i \}_{i \in \mathcal{A}}, V \rangle$  and a world $w\in W$ is expressed by writing that $\mathcal{M},w \models \varphi$.
We define $V^\mathcal{M}(\varphi) = \{w \in W \ | \ \mathcal{M},w \models \varphi \}$. Formula $\varphi$ is \textit{valid} if and only if for all $\mathcal{M}$ and for all worlds $w$ we have $\mathcal{M},w \models \varphi$.

%Modal logic K is 
Our modal logic above (corresponding to the normal modal logic K) is not yet sufficiently suited to encode epistemic reasoning. Therefore, additional conditions (reflexivity, transitivity and euclideaness) are imposed on the accessibility relations. 
This can e.g.~be achieved by postulating the following principles, resp.~axiom schemata (in a Hilbert-style proof system).
% $\mathcal{S}5$ describes the so-called \textit{properties of knowledge}. In $\mathcal{S}5$ the accessibility relations are equivalence classes.

\begin{center}
	\begin{tabular}{llll}
		& \textbf{Assumption} & \textbf{Formula} & \textbf{Property}\\
		\hline
		\textbf{T} & \textit{Truth} & $K_i \varphi \rightarrow \varphi$ & Reflexive\\
		%\textbf{D.} & \textit{Consistency} & $K_i \varphi \rightarrow \neg K_i \neg \varphi$ & Serial\\
		%\textbf{B.} & & $\varphi \rightarrow K_i M_i \varphi$ & symmetric\\
		\textbf{4} & \textit{Positive Introspection} & $K_i \varphi \rightarrow K_i K_i \varphi$ & Transitive\\
		\textbf{5} & \textit{Negative Introspection} & $\neg K_i \rightarrow K_i \neg K_i \varphi$ & Euclidean
	\end{tabular}
\end{center}

We add public announcements \cite{plaza1989} to our logic.
The objective is to formulate an operation that transforms the epistemic model such that all agents \textit{find out that $\varphi$ is true.}
This is achieved by taking the model $\mathcal{M}$ and discarding all worlds in which $\varphi$ is false.
Afterwards all agents will only consider worlds in which $\varphi$ is true.
Because of the \textit{publicity} of the announcement all agents are aware of the fact that all other agents know that $\varphi$ holds true afterwards.

\begin{definition}[\textbf{Public Announcement}]
	Suppose that $\mathcal{M} = \langle W, \{R_i\}_{i\in\mathcal{A}}, V \rangle$ is an epistemic model and $\varphi$ is a formula (in the language of our base logic).
		After all the agents find out that $\varphi$ is true (\textit{i.e., $\varphi$ is publicly announced}), the resulting model is
		$\mathcal{M}^{!\varphi} = \langle W^{!\varphi}, \{R_i^{!\varphi}\}_{i\in\mathcal{A}}, V^{!\varphi} \rangle$ where $W^{!\varphi} = \{ w \in W \ | \ \mathcal{M},w \models \varphi \}$,\\
		$R_i^{!\varphi} = R_i \cap (W^{!\varphi} \times W^{!\varphi})$ for all $i \in \mathcal{A}$, and $V^{!\varphi}(p) = V(p) \cap W^{!\varphi}$ for all $p \in \mathcal{P}$.
		
		To say that "\textit{$\psi$ is true after the announcement of $\varphi$}" is represented as $[!\varphi]\psi$.
		Truth for this new operator is defined as:
		%\[ \mathcal{M}, w \models [!\varphi]\psi \text{ iff } \mathcal{M},w \models \varphi \textrm{ then } \mathcal{M}^{!\varphi}, w \models \psi \]
		\[ \mathcal{M}, w \models [!\varphi]\psi \text{ iff } \mathcal{M},w \not\models \varphi \textrm{ or } \mathcal{M}^{!\varphi}, w \models \psi \]
		
\end{definition}

\noindent
We conclude this section with the introduction of notions for group knowledge.

Mutual knowledge, often stated as \textit{everyone knows}, describes knowledge that each member of the group holds.
Usually, it is defined for a group of agents $G \subseteq \mathcal{A}$ as $E_G \varphi := \bigwedge_{i \in G}K_i \varphi$.
Equivalently, a new relation can be introduced to express mutual knowledge with the knowledge operator.

\begin{definition}[\textbf{Mutual Knowledge}]
    Let $G \subseteq \mathcal{A}$ be  a group of agents.
	Let $R_G = \bigcup_{i \in G}R_i$.
	The truth clause for mutual knowledge is:
	\[
		\mathcal{M},w \models E_G \psi \text{ iff for all $v \in W$, if } wR_Gv \text{ then } \mathcal{M}, v \models \psi
	\]
\end{definition}

\noindent
Still, there is a distinction to make between \textit{everyone knows} $\varphi$ and \textit{it is common knowledge that} $\varphi$.
A statement $p$ is common knowledge when all agents know $p$, know that they all know $p$, know that they all know that they all know $p$, and so ad infinitum.
Relativized common knowledge was introduced by van Benthem, van Eijck and Kooi \cite{benthem2006} as a variant of common knowledge for dynamic epistemic logics.
As the name suggests knowledge update is then treated as a \textit{relativization}.
%Thus, public announcement logic with relativized common knowledge is reducible. \hl{Warum "Thus"? Was heisst "reducible"? Reducible to what?} With ordinary common knowledge that wouldn't be the case \cite{10.5555/645876.671885}.

\begin{definition}[\textbf{Relativized Common Knowledge}]
	Let $G \subseteq \mathcal{A}$ be  a group of agents.
	Let $R_G = \bigcup_{i \in G}R_i$.
	The truth clause for relativized common knowledge is:
	%\[\mathcal{M},w \models \mathcal{C}^\varphi_G \psi \text{ iff for all $v \in W$, if } w(R_G^\varphi)^+v \text{ then } \mathcal{M}, v \models \psi\]
	\vspace{-.25em}
	\[\mathcal{M},w \models \mathcal{C}_G(\varphi | \psi) \text{ iff for all $v \in W$, if } w(R_G^\varphi)^+v \text{ then } \mathcal{M}, v \models \psi\]
	where $R_G^\varphi = R_G \cap (W \times $\hl{$V^{\mathcal{M}}(\varphi)$}), and $(R_G^\varphi)^+$ denotes the transitive closure of $R_G^\varphi$.
\end{definition}

\noindent
%Intuitively, $\mathcal{C}_G^\varphi\psi$ expresses, that after $\varphi$ is announced, $\psi$ becomes common knowledge in the group.
Intuitively, $\mathcal{C}_G(\varphi | \psi)$ expresses, that after $\varphi$ is announced, $\psi$ becomes common knowledge in the group.
This means, that every path from $w$, that is accessible using the agent's relations through worlds in which $\varphi$ is true, must end in a world in which $\psi$ is true.
%We can still express ordinary common knowledge without its preceding logical sentence.
%Ordinary common knowledge of $\varphi$ can be abbreviated as $\mathcal{C}^\top_G\varphi$, where $\top$ denotes an arbitrary tautology.
Ordinary common knowledge of $\varphi$ can be abbreviated as $\mathcal{C}_G(\top | \varphi)$, where $\top$ denotes an arbitrary tautology.

In the remainder we use PAL to refer to the depicted logic consisting of \hl{modal logic K}, extended by the principles T45, public announcement and relativized common knowledge.

\section{Modeling PAL as a Fragment of HOL}

%Before continuing with providing an PAL into HOL, some words about the here employed \textit{shallow semantical embedding} (SSE) approach are in order.
A shallow semantical embedding (SSE) of a target logic into HOL provides a translation between  the two logics in such a way that the former logic is identified and characterized as a proper fragment of the latter.
%Syntax and semantic of the desired target logic are explicitly modeled in HOL, which acts as our meta-logic. 
%Leveraging this, we can translate proof problems of the target logic into a fragment of HOL.
Once such an SSE is obtained, all that is needed is to prove (or refute) conjectures in the target logic is to provide the SSE, encoded in an input file, to the HOL prover in addition to the encoded conjecture.
We can then use the HOL prover as-is, without making any changes to its source code, and use it to solve problems in our target logic.
% The effectiveness of the embedding approach has remarkably been demonstrated by formalizing G\"odels ontological argument \cite{C40}.

\vspace{-.45em}
\subsection{Shallow Semantical Embedding}
	
\vspace{-.15em}
To define an SSE for target logic PAL we lift the type of propositions in order to explicitly encode their dependency on possible worlds; this is analogous to prior work \cite{J41,J44}.
In order to capture the model-changing behavior of PAL we additionally introduce world domains (sets of worlds) as parameters/arguments in the encoding.
The rationale thereby is to be able to suitably constrain, and recursively pass-on, these domains after each model changing action.

PAL formulas are thus identified in our semantical embedding with certain HOL terms (predicates) of type \mbox{$(i \rightarrow o) \rightarrow i \rightarrow o$}. 
They can be applied to terms of type  \mbox{$i\rightarrow o$}, which are assumed to denote evaluation domains, and subsequently to terms of type $i$, which are assumed to denote possible worlds.
That is, the HOL type $i$ is identified with a (non-empty) set of worlds, and the type \mbox{$i \rightarrow o$}, abbreviated by $\sigma$, is identified with a set of sets of worlds, i.e., a set of evaluation domains.
%Consequently, while most semantical embeddings lift the propositions to \mbox{$i\rightarrow o$}, the here presented solution identifies (most) PAL terms of type \mbox{$(i\rightarrow o)\rightarrow i\rightarrow o$}.
%This allows us to check whether given worlds are part of the desired domain and consequently the truth of a formula can explicitly be evaluated.
%We introduce Type $\sigma$ as abbreviation for \mbox{$i \rightarrow o$}.
	Type \mbox{$(i \rightarrow o) \rightarrow i \rightarrow o$} is abbreviated as $\tau$, %\footnote{In similar work $\tau$ gets used as an abbreviation for \mbox{$i\rightarrow o$}. Still, $\tau$ is adapted here for the presented abbreviation in order to stay consistent with the literature.}.
	and type $\alpha$ is an abbreviation for \mbox{$i\rightarrow i\rightarrow o$}, the type of accessibility relations between worlds.
	
		For each propositional symbol $p^i$ of PAL, the corresponding HOL signature is assumed to contain a corresponding constant symbol $p^i_\sigma$, which is (rigidly) denoting the set of all those worlds in which $p^i$ holds.
		We call the $p^i_\sigma$ \textit{$\sigma$-type-lifted propositions}. 
Moreover, for $k = 1, \dots , |\mathcal{A}|$ the HOL signature is assumed to contain the constant symbols $r^1_\alpha, \dots, r^{|\mathcal{A}|}_\alpha$. 	Without loss of generality, we assume that besides those constants symbols and the primitive logical connectives of HOL, no other constant symbols are given in the signature of HOL.

	As a simplifying assumption in this ongoing work (which has a particular focus on an automation of the Wise Men Puzzle in PAL) we continue with choosing $|\mathcal{A}|=3$. (A generalization for arbitrary $\mathcal{A}$ is straightforward).

	The mapping $\lfloor \cdot \rfloor$ translates a formula $\varphi$ of PAL into a term $\lfloor \varphi \rfloor$ of HOL of type $\tau$.
	The mapping is defined recursively:
	\begin{align*}
	\lfloor p^j \rfloor &= (^A(p^j_\sigma))_\tau\\
	%\lfloor p^i \rfloor &= p_\tau^i\\
	\lfloor \neg \varphi \rfloor &= \neg_{\tau\rightarrow\tau} \lfloor \varphi \rfloor\\
	\lfloor \varphi \vee \psi \rfloor &= \vee_{\tau\rightarrow\tau\rightarrow\tau} \lfloor \varphi \rfloor \lfloor \psi \rfloor\\
	\lfloor K \ \text{r}^k \ \varphi \rfloor &= K_{\alpha \rightarrow \tau \rightarrow \tau} \ \text{r}^k_{\alpha} \ \lfloor \varphi \rfloor\\
	\lfloor [! \varphi ] \psi \rfloor &= [! \ \cdot \ ] \cdot_{\tau \rightarrow \tau \rightarrow \tau} \lfloor \varphi \rfloor \lfloor \psi \rfloor\\
	\lfloor \mathcal{C}( \varphi | \psi ) \rfloor &= \mathcal{C}( \cdot | \cdot)_{\tau \rightarrow \tau \rightarrow \tau} \lfloor \varphi \rfloor \lfloor \psi \rfloor
	\end{align*}
		Operator $^A(\cdot)$, which evaluates atomic formulas, is defined as follows:
   \begin{align*}
	^A\cdot_{\sigma\rightarrow\tau} &= \lambda A_\sigma \lambda D_\sigma \lambda X_i (D \ X \wedge A\ X)
	\end{align*}
		As a first argument it accepts a $\sigma$-type-lifted proposition $A_\sigma$, which are rigidly interpreted. As a second argument it accepts an evaluation domain $D_\sigma$, that is, an arbitrary subset of the domain associated with type $\sigma$. And as a third argument it accepts a current world. It then checks whether (i) the current world is a member of evaluation domain $D_\sigma$ and (ii) whether the $\sigma$-type-lifted proposition $A_\sigma$ holds in the current world.
		
		The other logical connectives of PAL, except for $[! \ \cdot \ ] \cdot_{\tau \rightarrow \tau \rightarrow \tau}$,  are now defined in a way so that they simply pass-on the evaluation domains as parameters to the atomic-level. Only $[! \ \cdot \ ] \cdot_{\tau \rightarrow \tau \rightarrow \tau}$ is modifying, in fact, constraining,  the evaluation domain it passes on, and it does this in the expected way (cf. Def.~3):
%
	%$^A\cdot_{\sigma\rightarrow\tau}$, $\neg_{\tau\rightarrow\tau}$, $\vee_{\tau\rightarrow\tau\rightarrow\tau}$, $K_{\alpha \rightarrow \tau \rightarrow \tau}$, and $\langle ! \ \cdot \ \rangle \cdot_{\tau \rightarrow \tau \rightarrow \tau}$ abbreviate the following terms of HOL:\footnote{The reader should not confuse $D_\sigma$, used to describe the domain of worlds, in the embedding with $\mathcal{D}_\sigma$ in the definition of Henkin semantics.}
%	
	\begin{align*}
	% ^A\cdot_{\sigma\rightarrow\tau} &= \lambda A_\sigma \lambda D_\sigma \lambda X_i (D \ X \wedge A\ X)\\
	\neg_{\tau\rightarrow\tau} &= \lambda A_\tau \lambda D_\sigma \lambda X_i \neg (A \ D \ X) \\ \vee_{\tau\rightarrow\tau\rightarrow\tau} &= \lambda A_\tau \lambda B_\tau \lambda D_\sigma \lambda X_i (A \ D \ X \vee B \ D \ X) \\
	\hl{\texttt{K}_{\alpha \rightarrow \tau \rightarrow \tau}} &\hl{= \lambda R_\alpha \lambda A_\tau \lambda D_\sigma \lambda X_i \forall Y_i ((D\ Y\ \wedge\ R \ X \ Y)\ \longrightarrow A \ D \ Y)}\\
	% \texttt{K}_{\alpha \rightarrow \tau \rightarrow \tau} &= \lambda R_\alpha \lambda A_\tau \lambda D_\sigma \lambda X_i \forall Y_i (\neg (D\ Y\ \wedge\ R \ X \ Y)\ \vee A \ D \ Y)\\
	[! \ \cdot \ ] \cdot_{\tau \rightarrow \tau \rightarrow \tau} &= \lambda A_\tau \lambda B_\tau \lambda D_\sigma \lambda X_i (\neg (A \ D \ X) \vee (B \ (\lambda Y_i\ D\ Y\ \wedge \ A\ D\ Y)\ X))
	\end{align*}
%
	%We also need to consider the case of atomic sentences.
	%This is necessary to differentiate between atomic sentences and other epistemic formulas, due to failures of uniform substitution. %TODO CITE
	%\[\lfloor ^A(p^j) \rfloor = {^A\cdot}_{\sigma\rightarrow\tau} \lfloor p^j \rfloor\]
%	
	To model $\mathcal{C}( \cdot | \cdot)_{\tau \rightarrow \tau \rightarrow \tau}$ we reuse the following operations on relations; cf.~\cite{J41,J44}.
	\begin{align*}
	\texttt{transitive}_{\alpha\rightarrow o} &= \lambda R_\alpha \forall X_i \forall Y_i \forall Z_i (\neg(R\ X\ Y\ \wedge\ R\ Y\ Z)\ \vee\ R\ X\ Z)\\
	\texttt{intersection}_{\alpha\rightarrow\alpha\rightarrow\alpha} &= \lambda R_\alpha \lambda Q_\alpha \lambda X_i \lambda Y_i (R\ X\ Y\ \wedge\ Q\ X\ Y)\\
	\texttt{union}_{\alpha\rightarrow\alpha\rightarrow\alpha} &= \lambda R_\alpha \lambda Q_\alpha \lambda X_i \lambda Y_i (R\ X\ Y\ \vee\ Q\ X\ Y)\\
	\texttt{sub}_{\alpha\rightarrow\alpha\rightarrow o} &= \lambda R_\alpha \lambda Q_\alpha \forall X_i \forall Y_i (\neg R\ X\ Y\ \vee\ Q\ X\ Y)\\
	\texttt{tc}_{\alpha\rightarrow\alpha} &= \lambda R_\alpha \lambda X_i \lambda Y_i \forall Q_\alpha\\
	&\hspace*{1.5em}(\neg \texttt{transitive}\ Q\ \vee\ (\neg \texttt{sub}\ R\ Q\ \vee\ Q\ X\ Y))
	% &\hspace*{1.5em}\hl{(\texttt{transitive}\ Q\ \longrightarrow (\texttt{sub}\ R\ Q\ \longrightarrow Q\ X\ Y))}
	\end{align*}
    Additionally, \texttt{EVR} is defined as the union of three agents $r^1, r^2$ and $r^3$ of type $\alpha$. \texttt{EVR} can then be used as a relation, e.g., for the knowledge operator to describe mutual knowledge of the three agents.
	But most importantly, we need this relation in order to encode relativized common knowledge.
%	
	%Additionally, \texttt{EVR} is defined as the union of three agents $r^1, r^2$ and $r^3$ of type $\alpha$. 
	%We need this relation in order to translate relativized common knowledge.
	\[\texttt{EVR}_\alpha = \texttt{union} (\texttt{union}\ r^1\ r^2)\ r^3\]
We want to remark that a general higher-order definition for the union of a set of relations could alternatively be introduced first and then be applied to our concrete set of relations $R$ consisting of $r^1$, $r^2$ and $r^3$. Nothing prevents us from generalizing the notion of mutual knowledge this way to an arbitrary  group of agents $R$, and to consider $R$ as a further parameter in e.g.~the definition of $\mathcal{C}( \cdot | \cdot)_{\tau \rightarrow \tau \rightarrow \tau}$.  However, in our first experiments as presented in this student paper, which are primarily intended to study the practical feasibility of the embedding approach for PAL, we have still avoided this final generalization step. The operator $\mathcal{C}( \cdot | \cdot)_{\tau \rightarrow \tau \rightarrow \tau}$ thus abbreviates the following HOL term:
%	
%	\begin{align*}
%	 \mathcal{C}( \cdot | \cdot)_{\tau \rightarrow \tau \rightarrow \tau} &= \lambda A_\tau \lambda B_\tau \lambda D_\sigma \lambda X_i \forall Y_i\\
%	&\hspace*{1.5em}(\neg (\texttt{tc} (\texttt{intersection} \ \texttt{EVR}\ (\lambda U_i \lambda V_i (D\ V\ \wedge\ A\ D\ V))))\ X\ Y\\
%	&\hspace*{1.5em}\vee\ B\ D\ Y)   
%	\end{align*}
%	
	\begin{align*}
	 \hl{\mathcal{C}( \cdot | \cdot)_{\tau \rightarrow \tau \rightarrow \tau}} &\hl{= \lambda A_\tau \lambda B_\tau \lambda D_\sigma \lambda X_i \forall Y_i}\\
	&\hspace*{1.5em}\hl{(\texttt{tc} (\texttt{intersection} \ \texttt{EVR}\ (\lambda U_i \lambda V_i (D\ V\ \wedge\ A\ D\ V)))\ X\ Y}\\
	&\hspace*{1.75em}\hl{\longrightarrow \ B\ D\ Y)}   
	\end{align*}
	Analyzing the truth of a PAL formula $\varphi$, represented by the HOL term $\lfloor \varphi \rfloor$, in a particular domain $d$, represented by the term $D_\sigma$, and a world $s$, represented by the term $S_i$, corresponds to evaluating the application ($\lfloor \varphi \rfloor \ D_\sigma \ S_i$).
	%For the embedding of the chosen approach we need also to keep an eye on the domain $D_\sigma$.
	%We verify whether $S$ denotes in $D$ by checking if $(D\ S)$ is true.
	%If that is the case, we evaluate $\varphi$ for this domain and world. 
%	
%	$\varphi$ is thus generally valid if and only if for all  $D_\sigma$ and all  $S_i$ we have $\neg D\ S \vee \lfloor \varphi \rfloor D\ S$.
	$\varphi$ is thus generally valid if and only if for all  $D_\sigma$ and all  $S_i$ we have \hl{$D\ S \rightarrow \lfloor \varphi \rfloor D\ S$}.

	The validity function, therefore, is defined as follows:
	% \[ \texttt{vld}_{\tau \rightarrow o} = \lambda A_\tau \forall D_\sigma \forall S_i (\neg D\ S\ \vee\ A \ D\ S).\]
	\[ \hl{\texttt{vld}_{\tau \rightarrow o} = \lambda A_\tau \forall D_\sigma \forall S_i (D\ S\ \longrightarrow\ A \ D\ S).}\]

The necessity to quantify over all possible domains in this definition will be further illustrated below.

\subsection{Encoding into Isabelle/HOL}	

What follows is a description of the concrete encoding of the presented SSE of PAL in HOL within the higher-order proof assistant Isabelle/HOL.\footnote{\hl{The full sources of our encoding can be found at \url{http://logikey.org} in subfolder \texttt{Public-Announcement-Logic}, resp.~at \url{https://github.com/cbenzmueller/LogiKEy/tree/master/Public-Announcement-Logic}.}}

All necessary types can be modeled in a straightforward way.
We declare \texttt{i} to denote possible worlds and then introduce type aliases for $\sigma$, $\tau$ and $\alpha$.
Type \texttt{bool} represents (the bivalent set of) truth values.

%\begin{lstlisting}
%typedecl i
%type_synonym $\sigma$ = "i$\Rightarrow$bool"
%type_synonym $\tau$ = "$\sigma\Rightarrow$i$\Rightarrow$bool"
%type_synonym $\alpha$ = "i$\Rightarrow$i$\Rightarrow$bool"
%\end{lstlisting}

\vspace{.5em}
\noindent
\lstinline[language=Isabelle]|typedecl i|\\
\lstinline[language=Isabelle]|type_synonym $\sigma$ = "i$\Rightarrow$bool"|\\
\lstinline[language=Isabelle]|type_synonym $\tau$ = "$\sigma\Rightarrow$i$\Rightarrow$bool"|\\
\lstinline[language=Isabelle]|type_synonym $\alpha$ = "i$\Rightarrow$i$\Rightarrow$bool"|
\vspace{.5em}

\noindent
The agents are declared mutually distinct accessibility relations and the group of agents is denoted by predicate $\mathcal{A}$. %\lstinline[language=Isabelle]|Agent|.
In order to obtain $\mathcal{S}5$ (KT45) properties, we declare respective conditions on the accessibility relations in the group of agents $\mathcal{A}$.
Various Isabelle/HOL encodings from \cite{J41,J44} are reused here (without mentioning due to space restrictions), including the encoding of transitive closure.

\vspace{.5em}
\noindent
\lstinline[language=Isabelle]|consts a::"$\alpha$$\textcolor{red}{"}$ b::"$\alpha$" c::"$\alpha$"|\\
\lstinline[language=Isabelle]|abbreviation "$\mathcal{A}$ x $\equiv$ x = a $\vee$ x = b $\vee$ x = c"|\\
\lstinline[language=Isabelle]|axiomatization where|\\
\hspace*{1.5em}\lstinline[language=Isabelle]|alldifferent: "$\neg(\texttt{a}=\texttt{b}) \wedge \neg(\texttt{a}=\texttt{c}) \wedge \neg(\texttt{b}=\texttt{c})$" and|\\
\hspace*{1.5em}\lstinline[language=Isabelle]|agents_S5: "$\forall$i. $\mathcal{A}$ i $\longrightarrow$ (reflexive i $\wedge$ transitive i $\wedge$ euclidean i)"|\\
\lstinline[language=Isabelle]|abbreviation EVR :: "$\alpha$" ("EVR")|\\
\hspace*{1.5em}\lstinline[language=Isabelle]|where "EVR $\equiv$ union_rel (union_rel a b) c"|
\vspace{.5em}

%\begin{lstlisting}
%consts a::"$\alpha$ b::"$\alpha$" c::"$\alpha$"
%abbreviation Agent ("$\mathcal{A}$") where
%  "$\mathcal{A}$ x $\equiv$ x = a $\vee$ x = b $\vee$ x = c"
%axiomatization where
%  alldifferent: "$\neg(\texttt{a}=\texttt{b}) \wedge \neg(\texttt{a}=\texttt{c}) \wedge \neg(\texttt{b}=\texttt{c})$
%  agents_S5: "$\forall$i. $\mathcal{A}$ i $\longrightarrow$ (reflexive i $\wedge$ transitive i
%                      $\wedge$ euclidean i)"
%definition EVR :: "$\alpha$" ("EVR")
%  where "EVR $\equiv$ union_rel (union_rel a b) c"
%\end{lstlisting}

\noindent
%Terms that can be constructed using $\neg$ or $\vee$ are lifted in a straightforward manner.
To distinguish between HOL connectives (e.g.~$\neg$) and the lifted PAL connectives (e.g.~$\boldsymbol{\neg}_{\tau \rightarrow \tau}$) we make use of bold face fonts, see for example the definition $\boldsymbol{\neg}_{\tau \rightarrow \tau} \equiv \lambda \varphi_\tau. \lambda W_{\sigma}. \lambda w_i. \neg \varphi\ \texttt{W w}$ below.
Each of the lifted unary and binary connectives of PAL accepts arguments of type $\tau$, i.e.~lifted PAL formulas, and returns such a lifted PAL formula.

A special case, as discussed before, is the new operator for atomic propositions $^A(\cdot)$.
When evaluating $\sigma$-type lifted atomic propositions $p$ we need to check if $p$ is true in the given world \texttt{w}, but we also need to check whether the given world \texttt{w} is still part of our evaluation domain \texttt{W} that has been recursively passed-on.
Operator $^A(\cdot)$ is thus of type "$\sigma \Rightarrow \tau$". %and thus allows to evaluate the given term dependent on the world under consideration.

\vspace{.5em}
\noindent	
\lstinline[language=Isabelle]|abbreviation patom :: "$\sigma\Rightarrow\tau$" ("$^\texttt{A}$_")|\\
\hspace*{1.5em}\lstinline[language=Isabelle]|where "$^\texttt{A}$p $\equiv$ $\lambda$W w. W w $\wedge$ p w"|\\
\lstinline[language=Isabelle]|abbreviation ptop :: $\tau$ ("$\boldsymbol{\top}$")|\\
\hspace*{1.5em}\lstinline[language=Isabelle]|where "$\boldsymbol{\top}$ $\equiv$ $\lambda$W w. True"|\\
\lstinline[language=Isabelle]|abbreviation pneg :: "$\tau$$\Rightarrow$$\tau$" ("$\boldsymbol{\neg}$")|\\
\hspace*{1.5em}\lstinline[language=Isabelle]|where "$\boldsymbol{\neg} \varphi$ $\equiv$ $\lambda$W w. $\neg(\varphi$ W w)"|\\
\lstinline[language=Isabelle]|abbreviation pand :: "$\tau$$\Rightarrow$$\tau$$\Rightarrow$$\tau$" ("$\boldsymbol{\wedge}$")|\\
\hspace*{1.5em}\lstinline[language=Isabelle]|where "$\varphi \boldsymbol{\wedge} \psi$ $\equiv$ $\lambda$W w. ($\varphi$ W w) $\wedge$ ($\psi$ W w)"|\\
\lstinline[language=Isabelle]|abbreviation por :: "$\tau$$\Rightarrow$$\tau$$\Rightarrow$$\tau$" ("$\boldsymbol{\vee}$")|\\
\hspace*{1.5em}\lstinline[language=Isabelle]|where "$\varphi \boldsymbol{\vee} \psi$ $\equiv$ 	$\lambda$W w. ($\varphi$ W w) $\vee$ ($\psi$ W w)"|\\
\lstinline[language=Isabelle]|abbreviation pimp :: "$\tau$$\Rightarrow$$\tau$$\Rightarrow$$\tau$" ("$\boldsymbol{\rightarrow}$")|\\
\hspace*{1.5em}\lstinline[language=Isabelle]|where "$\varphi \boldsymbol{\rightarrow} \psi$ $\equiv$ 	$\lambda$W w. ($\varphi$ W w) $\longrightarrow$ ($\psi$ W w)"|\\	
\lstinline[language=Isabelle]|abbreviation pequ :: "$\tau$$\Rightarrow$$\tau$$\Rightarrow$$\tau$" ("$\boldsymbol{\leftrightarrow}$")|\\
\hspace*{1.5em}\lstinline[language=Isabelle]|where "$\varphi \boldsymbol{\leftrightarrow} \psi$ $\equiv$ 	$\lambda$W w. ($\varphi$ W w) $\longleftrightarrow$ ($\psi$ W w)"|
\vspace{.5em}

%\begin{lstlisting}
%abbreviation ptop :: $\tau$ ("$\boldsymbol{\top}$")
%  where "$\boldsymbol{\top}$ $\equiv$ $\lambda$W w. True"
%abbreviation pneg :: "$\tau$$\Rightarrow$$\tau$" ("$\boldsymbol{\neg}$")
%  where $\boldsymbol{\neg} \varphi$ $\equiv$ $\lambda$W w. $\neg(\varphi$ W w) "
%abbreviation pand :: "$\tau$$\Rightarrow$$\tau$$\Rightarrow$$\tau$" ("$\boldsymbol{\wedge}$")
%  where $\varphi \boldsymbol{\wedge} \psi$ $\equiv$ $\lambda$W w. ($\varphi$ W w) $\wedge$ ($\psi$ W w) "
%abbreviation por :: "$\tau$$\Rightarrow$$\tau$$\Rightarrow$$\tau$" ("$\boldsymbol{\vee}$")
%  where $\varphi \boldsymbol{\vee} \psi$ $\equiv$ 	$\lambda$W w. ($\varphi$ W w) $\vee$ ($\psi$ W w) "
%abbreviation pimp :: "$\tau$$\Rightarrow$$\tau$$\Rightarrow$$\tau$" ("$\boldsymbol{\rightarrow}$")
%  where $\varphi \boldsymbol{\rightarrow} \psi$ $\equiv$ 	$\lambda$W w. ($\varphi$ W w) $\longrightarrow$ ($\psi$ W w) "
%abbreviation pequ :: "$\tau$$\Rightarrow$$\tau$$\Rightarrow$$\tau$" ("$\boldsymbol{\leftrightarrow}$")
%  where $\varphi \boldsymbol{\leftrightarrow} \psi$ $\equiv$ 	$\lambda$W w. ($\varphi$ W w) $\longleftrightarrow$ ($\psi$ W w)"
%\end{lstlisting}
\noindent
In the definition of the knowledge operator \texttt{K}, we have to make sure to add a domain check in the implication.
%The reasoning is the same as described when introducing $\texttt{vld}_{\tau \rightarrow \sigma}$ in the previous section.

\vspace{.5em}
\noindent \marginpar{\hl{modified}}
\lstinline[language=Isabelle]|abbreviation pknow :: "$\tau$$\Rightarrow$$\tau$$\Rightarrow$$\tau$" ("$\textbf{K}$ _ _")|\\
%\hspace*{1.5em}\lstinline[language=Isabelle]|where "$\textbf{K}$ r $\varphi$ $\equiv \lambda$W w.$\forall$v.$\textcolor{red}{\texttt{(W v $\wedge$ r w v) $\longrightarrow$ ($\varphi$ W v)}"}
 \hspace*{1.5em}\lstinline[language=Isabelle]|where "$\textbf{K}$ r $\varphi$ $\equiv \lambda$W w.$\forall$v. (W v $\wedge$ r w v)  $\longrightarrow$  ($\varphi$ W v)"|
% (| am Ende wieder einfuegen)
% \hspace*{1.5em}\lstinline[language=Isabelle]|where "$\textbf{K}$ r $\varphi$ $\equiv \lambda$W w.$\forall$v. $\neg$(W v $\wedge$ r w v) $\vee$ ($\varphi$ W v)"|
\vspace{.5em}

\noindent
Two additional abbreviations are introduced to improve readability.
A more concise way to state knowledge and an additional operator for mutual knowledge, in which the \texttt{EVR} relation gets used.

\vspace{.5em}
\noindent
\lstinline[language=Isabelle]|abbreviation agtknows :: "$\tau$$\Rightarrow$$\tau$$\Rightarrow$$\tau$" ("$\textbf{K}_{\_}$ _")|\\
\hspace*{1.5em}\lstinline[language=Isabelle]|where "$\textbf{K}_\texttt{r}\, \varphi$ $\equiv \textbf{K}$ r $\varphi$"|\\
\lstinline[language=Isabelle]|abbreviation evrknows :: "$\tau$$\Rightarrow$$\tau$" ("$\textbf{E}_\mathcal{A}$ _")|\\
\hspace*{1.5em}\lstinline[language=Isabelle]|where "$\textbf{E}_\mathcal{A}\, \varphi \equiv \textbf{K}$ EVR $\varphi$"|
\vspace{.5em}

%\begin{lstlisting}
%abbreviation pknow :: "$\tau$$\Rightarrow$$\tau$$\Rightarrow$$\tau$" ("$\pmb{K}$ _ _")
%  where $\pmb{K}$ r $\varphi$ $\equiv \lambda$W w.$\forall$v. (W v $\wedge$ r w v) $\longrightarrow$ ($\varphi$ W v)"
%\end{lstlisting}

\noindent
We finally see the change of the evaluation domain in action, when introducing the public announcement operator.
We already inserted domain checks in the definition of the operators \texttt{K} and $^A(\cdot)$.
Now\hl{,} we need to constrain the domain after each public announcement.
So far the evaluation domain, modeled by \texttt{W}, got passed-on through all lifted operators without any change.
In the public announcement operator, however, we modify the evaluation domain \texttt{W} into $(\lambda\texttt{z}.\ \texttt{W z}\ \wedge\ \varphi\ \texttt{W z})$ (i.e., the set of all worlds \texttt{z} in \texttt{W}, such that $\varphi$ holds for \texttt{W} and \texttt{z}),
which is then  recursively passed-on. 
The public announcement operator is thus defined as:

\vspace{.5em}
\noindent
\lstinline[language=Isabelle]|abbreviation ppal :: "$\tau$$\Rightarrow$$\tau$$\Rightarrow$$\tau$" ("$\boldsymbol{\text{[}}\boldsymbol{\text{!}}$_$\boldsymbol{\text{]}}$_")|\\
\hspace*{1.5em}\lstinline[language=Isabelle]|where "$\boldsymbol{\text{[}}\boldsymbol{\text{!}}\varphi\boldsymbol{\text{]}}\psi$ $\equiv$ $\lambda$W w. $\neg (\varphi$ W w) $\vee$ ($\psi$ ($\lambda$z. W z $\wedge$ $\varphi$ W z) w)"|
\vspace{.5em}

%\begin{lstlisting}
%abbreviation ppal :: "$\tau$$\Rightarrow$$\tau$$\Rightarrow$$\tau$" %("$\boldsymbol{\text{[}}\boldsymbol{\text{!}}$_$\boldsymbol{\text{]}}$_")
%  where $\boldsymbol{\text{[}}\boldsymbol{\text{!}}\varphi\boldsymbol{\text{]}}\psi$ $\equiv$ $\lambda$W w. $(\varphi$ W w) $\longrightarrow$ ($\psi$ ($\lambda$z. W z $\wedge$ $\varphi$ W z) w)"
%\end{lstlisting}
\noindent
The following embedding of relativized common knowledge is a straightforward encoding of the semantic properties and definitions as proposed in Def.~5.
%Other than that, the operator behaves exactly the same as the $\pmb{K}$ operator.

\vspace{.5em}
\noindent
\lstinline[language=Isabelle]|abbreviation prck :: "$\tau$$\Rightarrow$$\tau$$\Rightarrow\tau$" ("$\textbf{C}\boldsymbol{\llparenthesis}$_$\boldsymbol{|}$_$\boldsymbol{\rrparenthesis}$")|\\
\hspace*{1.5em}\lstinline[language=Isabelle]|where "$\textbf{C}\boldsymbol{\llparenthesis}\varphi\boldsymbol{|}\psi\boldsymbol{\rrparenthesis}$" $\equiv$ $\lambda$W w. $\forall$v.|\\
% \hspace*{3.5em}\lstinline[language=Isabelle]|$\neg$(tc (intersection_rel EVR ($\lambda$u v. W v $\wedge\ \varphi$ W w)) w v) $\vee$ ($\psi$ W v)"|
\hspace*{3.5em}\lstinline[language=Isabelle]|(tc (intersection_rel EVR ($\lambda$u v. W v $\wedge\ \varphi$ W w)) w v) $\textcolor{red}{\longrightarrow}$ ($\psi$ W v)"|
\vspace{.5em}

%\begin{lstlisting}
%abbreviation pccmn :: "$\tau$$\Rightarrow$$\tau$$\Rightarrow\tau$" ("$\textbf{C}\boldsymbol{\llparenthesis}$_$\boldsymbol{|}$_$\boldsymbol{\rrparenthesis}$")
%  where "$\textbf{C}\boldsymbol{\llparenthesis}\varphi\boldsymbol{|}\psi\boldsymbol{\rrparenthesis}$" $\equiv$ $\lambda$W w. $\forall$v. (tc(intersection_rel EVR
%                      ($\lambda$u v. W v $\wedge\ \varphi$ W w)) w v $\longrightarrow$ ($\psi$ W v))"
%\end{lstlisting}
\noindent
As described earlier we can abbreviate ordinary common knowledge as $\mathcal{C}_G(\top | \varphi)$:

\vspace{.5em}
\noindent
\lstinline[language=Isabelle]|abbreviation pcmn :: "$\tau$$\Rightarrow$$\tau$" ("$\textbf{C}_{\mathcal{A}}$ _") where "$\textbf{C}_{\mathcal{A}}\ \varphi$ $\equiv$ $\textbf{C}\boldsymbol{\llparenthesis}\boldsymbol{\top|}\varphi\boldsymbol{\rrparenthesis}$"|

\vspace{.5em}
\noindent
Finally an embedding for the notion of validity is needed.
Generally, for a type-lifted formula $\varphi$ to be valid, the application of $\varphi$ \sout{\hl{to \texttt{w}}} has to hold true for all worlds \texttt{w}. In the context of PAL the evaluation domains also have to be incorporated in the definition.
Originally we were tempted to define PAL validity in such that we start with a "full evaluation domain", a domain that evaluates to \texttt{True} for all possible worlds and gets restricted, whenever necessary after an announcement.
Such a validity definition would look like this:

\vspace{.5em}
\noindent
\lstinline[language=Isabelle]|abbreviation tvalid::"$\tau$$\Rightarrow$bool" ("$\lfloor$_$\rfloor^\texttt{T}$") where "$\lfloor$_$\rfloor^\texttt{T}$ $\equiv$ $\forall$w. $\varphi$ ($\lambda$x. True) w"|
\vspace{.5em}

%\begin{lstlisting}
%abbreviation tvalid :: "$\tau$$\Rightarrow$bool" ("$\lfloor$_$\rfloor^\texttt{T}$")
%  where $\lfloor$_$\rfloor^\texttt{T}$ $\equiv$ $\forall$w. $\varphi$ ($\lambda$x. True) w"
%\end{lstlisting}
\noindent
But this leads to undesired behavior, which we can easily see when using our reasoning tools to study e.g.~the validity of an often proposed schematic axiom of PAL,
 \textit{Announcement Necessitation}: $\textit{from}\ \psi\textit{, infer}\ [!\varphi]\psi$.
If we check for a counterexample in Isabelle/HOL, the model finder Nitpick reports the following:

\vspace{.5em}
\noindent
\lstinline[language=Isabelle]|lemma necessitation: assumes "$\lfloor \psi \rfloor^\texttt{T}$" shows "$\lfloor\boldsymbol{[!}\varphi\boldsymbol{]}\psi\rfloor^\texttt{T}$" nitpick oops|\\
\noindent\rule{\textwidth}{0.4pt}

\noindent
\lstinline[language=Isabelle]|Nitpick found a counterexample for card i = 2:|

\noindent
\lstinline[language=Isabelle]|Free variables:|

\noindent
\lstinline[language=Isabelle]|$\varphi$ = ($\lambda$x. _)|\\
\hspace*{2em}\lstinline[language=Isabelle]|((($\lambda$x. _)($\texttt{i}_1$ := True, $\texttt{i}_2$ := True), $\texttt{i}_1$) := False,|\\
\hspace*{2.5em}\lstinline[language=Isabelle]|(($\lambda$x. _)($\texttt{i}_1$ := True, $\texttt{i}_2$ := True), $\texttt{i}_2$) := True,|\\
\hspace*{2.5em}\lstinline[language=Isabelle]|(($\lambda$x. _)($\texttt{i}_1$ := True, $\texttt{i}_2$ := False), $\texttt{i}_1$) := False,|\\
\hspace*{2.5em}\lstinline[language=Isabelle]|(($\lambda$x. _)($\texttt{i}_1$ := True, $\texttt{i}_2$ := False), $\texttt{i}_2$) := False,|\\
\hspace*{2.5em}\lstinline[language=Isabelle]|(($\lambda$x. _)($\texttt{i}_1$ := False, $\texttt{i}_2$ := True), $\texttt{i}_1$) := False,|\\
\hspace*{2.5em}\lstinline[language=Isabelle]|(($\lambda$x. _)($\texttt{i}_1$ := False, $\texttt{i}_2$ := True), $\texttt{i}_2$) := False,|\\
\hspace*{2.5em}\lstinline[language=Isabelle]|(($\lambda$x. _)($\texttt{i}_1$ := False, $\texttt{i}_2$ := False), $\texttt{i}_1$) := False,|\\
\hspace*{2.5em}\lstinline[language=Isabelle]|(($\lambda$x. _)($\texttt{i}_1$ := False, $\texttt{i}_2$ := False), $\texttt{i}_2$) := False)|

\noindent
\lstinline[language=Isabelle]|$\psi$ = ($\lambda$x. _)|\\
\hspace*{2em}\lstinline[language=Isabelle]|((($\lambda$x. _)($\texttt{i}_1$ := True, $\texttt{i}_2$ := True), $\texttt{i}_1$) := True,|\\
\hspace*{2.5em}\lstinline[language=Isabelle]|(($\lambda$x. _)($\texttt{i}_1$ := True, $\texttt{i}_2$ := True), $\texttt{i}_2$) := True,|\\
\hspace*{2.5em}\lstinline[language=Isabelle]|(($\lambda$x. _)($\texttt{i}_1$ := True, $\texttt{i}_2$ := False), $\texttt{i}_1$) := False,|\\
\hspace*{2.5em}\lstinline[language=Isabelle]|(($\lambda$x. _)($\texttt{i}_1$ := True, $\texttt{i}_2$ := False), $\texttt{i}_2$) := False,|\\
\hspace*{2.5em}\lstinline[language=Isabelle]|(($\lambda$x. _)($\texttt{i}_1$ := False, $\texttt{i}_2$ := True), $\texttt{i}_1$) := False,|\\
\hspace*{2.5em}\lstinline[language=Isabelle]|(($\lambda$x. _)($\texttt{i}_1$ := False, $\texttt{i}_2$ := True), $\texttt{i}_2$) := False,|\\
\hspace*{2.5em}\lstinline[language=Isabelle]|(($\lambda$x. _)($\texttt{i}_1$ := False, $\texttt{i}_2$ := False), $\texttt{i}_1$) := False,|\\
\hspace*{2.5em}\lstinline[language=Isabelle]|(($\lambda$x. _)($\texttt{i}_1$ := False, $\texttt{i}_2$ := False), $\texttt{i}_2$) := False)|\\
\lstinline[language=Isabelle]|Skolem constant:|\\
\hspace*{2em}\lstinline[language=Isabelle]|??.tvalid.w = $\texttt{i}_2$|\\

%\noindent\rule{\textwidth}{0.4pt}

%Here, we can see that in world $\texttt{i}_2$ the term $\lfloor \psi \rfloor^\texttt{T}$  is true.
%Yet, when we want to evaluate $\lfloor \boldsymbol{[!}\varphi\boldsymbol{]}\psi \rfloor^\texttt{T}$ the term is false.
%This is because with announcing $\psi$ we discard all worlds in which $\psi$ does not hold.
%Thus, ultimately the term evaluates as follows: 
%$\texttt{(($\lambda$x. \_)($\texttt{i}_1$ := False, $\texttt{i}_2$ := True), $\texttt{i}_2$) := False}$
\noindent
 The valid function needs instead to be defined such that it checks validity not only for all worlds, but for all domains and worlds. Otherwise, the observed but undesired value flipping may occur.

\vspace{.5em}
\noindent
\lstinline[language=Isabelle]|abbreviation pvalid :: "$\tau$$\Rightarrow$bool" ("$\lfloor$_$\rfloor$")|\\
\hspace*{1.5em}\lstinline[language=Isabelle]|where "$\lfloor$_$\rfloor$ $\equiv$ $\forall$W.$\forall$w. W w $\longrightarrow$ $\varphi$ W w "|
\vspace{.5em}

%\begin{lstlisting}
%abbreviation pvalid :: "$\tau$$\Rightarrow$bool" ("$\lfloor$_$\rfloor$")
%  where $\lfloor$_$\rfloor$ $\equiv$ $\forall$W.$\forall$w. W w $\longrightarrow$ $\varphi$ W w 
%\end{lstlisting}
\noindent
All here introduced definitions are hidden from the user, who can construct formulas in PAL and prove these using the newly embedded operators.

\section{Experiments}
\subsection{Proving Axioms and Rules of Inference of PAL in HOL}

The presented SSE of PAL is able to prove the following axioms and rules of inference as presented for PAL in 
%the \textit{axiomatic theories of relativized common knowledge} 
\cite[see also Appendix F]{sep-dynamic-epistemic}:

\vspace{.5em}
\noindent
\begin{tabular}{p{10em}l}
\multicolumn{2}{l}{\textbf{System K}}\\
-- & All substitutions instances of propositional tautologies\\
Axiom K & $K_i(\varphi \rightarrow \psi) \rightarrow (K_i \varphi \rightarrow K_i \psi)$\\
Modus ponens & From $\varphi$ and $\varphi \rightarrow \psi$ infer $\psi$\\
Necessitation & From $\varphi$ infer $K_i \varphi$\\
\end{tabular}

\noindent
\begin{tabular}{p{10em}l}
\multicolumn{2}{l}{\textbf{System $\mathcal{S}5$}}\\
Axiom T & $K_i \varphi \rightarrow \varphi$\\
%Axiom D & $M_i \varphi$\\
%Axiom B & $\varphi \rightarrow K_i M_i \varphi$\\
Axiom 4 & $K_i \varphi \rightarrow K_i K_i \varphi$\\
Axiom 5 & $\neg K_i \varphi \rightarrow K_i \neg K_i \varphi$\\
\end{tabular}

\noindent
\begin{tabular}{p{10em}l}
\multicolumn{2}{l}{\textbf{Reduction Axioms}}\\
Atomic Permanence & $[!\varphi]p \leftrightarrow (\varphi \rightarrow p)$\\
Conjunction & $[!\varphi](\psi \wedge \chi) \leftrightarrow ([!\varphi]\psi \wedge [!\varphi]\chi)$\\
Partial Functionality & $[!\varphi]\neg \psi \leftrightarrow (\varphi \rightarrow \neg [!\varphi]\psi)$\\
Action-Knowledge & $[!\varphi]K_i\psi \leftrightarrow (\varphi \rightarrow K_i (\varphi \rightarrow K_i (\varphi \rightarrow [!\varphi]\psi)))$\\
-- & $[!\varphi]\mathcal{C}(\chi | \psi) \leftrightarrow (\varphi \rightarrow \mathcal{C}(\varphi \wedge [!\varphi]\chi | [!\varphi]\psi))$\\
\end{tabular}

\noindent
\begin{tabular}{p{10em}l}
\multicolumn{2}{l}{\textbf{Axiom schemes for RCK}}\\
$\mathcal{C}$-normality & $\mathcal{C}(\chi | (\varphi \rightarrow \psi)) \rightarrow (\mathcal{C}(\chi | \varphi) \rightarrow \mathcal{C}(\chi | \psi))$\\
Mix axiom & $\mathcal{C}(\psi | \varphi) \leftrightarrow E(\psi \rightarrow (\varphi \wedge \mathcal{C}(\psi | \varphi)))$\\
Induction axiom & $(E(\psi \rightarrow \varphi) \wedge \mathcal{C}(\psi | \varphi \rightarrow E(\psi \rightarrow \varphi))) \rightarrow \mathcal{C}(\psi | \varphi)$
\end{tabular}

\noindent
\begin{tabular}{p{10em}l}
\multicolumn{2}{l}{\textbf{Rules of Inference}}\\
Announcement Nec. & from $\varphi$, infer $[!\psi]\varphi$\\
RKC Necessitation & from $\varphi$, infer $\mathcal{C}(\psi | \varphi)$
\end{tabular}
\vspace{.25em}

\noindent
Only for the mix- and induction axiom (schemata) for relativized common knowledge is one direction, respectively, not automatically provable yet. Structural induction is required and a proof still needs to be provided by hand.
%For the \textit{RKC Necessitation}, both provers \textit{vampire} and \textit{e} suggested the stated proof using metis.
%Unfortunately, Isabelle was not able to reconstruct this proof.

\noindent
\lstinline[language=Isabelle]|$\textcolor{gray}{\text{(*System K*)}}$|\\
\lstinline[language=Isabelle]|lemma tautologies: "$\boldsymbol{\lfloor\top\rfloor}$" by auto|\\
%\hspace*{1.5em}\lstinline[language=Isabelle]|by auto|\\
\lstinline[language=Isabelle]|lemma axiom_K: "$\mathcal{A}$ i $\Longrightarrow\boldsymbol{\lfloor}(\textbf{K}_\texttt{i}\ (\varphi\boldsymbol{\rightarrow}\psi))\boldsymbol{\rightarrow}((\textbf{K}_\texttt{i}\ \varphi)\boldsymbol{\rightarrow}(\textbf{K}_\texttt{i}\ \psi))\boldsymbol{\rfloor}$" by auto|\\
%\hspace*{1.5em}\lstinline[language=Isabelle]|by auto|\\
\lstinline[language=Isabelle]|lemma modusponens: assumes 1: "$\boldsymbol{\lfloor}\varphi \boldsymbol{\rightarrow} \psi\boldsymbol{\rfloor}$" and 2: "$\boldsymbol{\lfloor}\varphi\boldsymbol{\rfloor}$" shows "$\boldsymbol{\lfloor}\psi\boldsymbol{\rfloor}$"|\\
\hspace*{1.5em}\lstinline[language=Isabelle]|using 1 2 by auto|\\
\lstinline[language=Isabelle]|lemma necessitation: assumes 1: "$\boldsymbol{\lfloor}\varphi\boldsymbol{\rfloor}$" shows "$\mathcal{A}$ i $\Longrightarrow \boldsymbol{\lfloor} \textbf{K}_\texttt{i}\ \varphi\boldsymbol{\rfloor}$"|\\
\hspace*{1.5em}\lstinline[language=Isabelle]|using 1 by auto|\\
\lstinline[language=Isabelle]|$\textcolor{gray}{\text{(*More axiom systems*)}}$|\\
\lstinline[language=Isabelle]|lemma axiom_T: "$\mathcal{A}$ i $\Longrightarrow \boldsymbol{\lfloor}(\textbf{K}_\texttt{i}\ \varphi) \boldsymbol{\rightarrow} \varphi) \boldsymbol{\rfloor}$"|\\
\hspace*{1.5em}\lstinline[language=Isabelle]|using group_S5 reflexive_def by auto|\\
%\lstinline[language=Isabelle]|lemma axiom_D: "$\mathcal{A}$ i $\Longrightarrow \boldsymbol{\lfloor}\textbf{M}_\texttt{i}\ \boldsymbol{\top} \boldsymbol{\rfloor}$"|\\
%\hspace*{1.5em}\lstinline[language=Isabelle]|using group_S5 pknow_def pneg_def pvalid_def reflexive_def tautologies by auto|
%\lstinline[language=Isabelle]|lemma axiom_B: "$\mathcal{A}$ i $\Longrightarrow \boldsymbol{\lfloor}\varphi \boldsymbol{\rightarrow}(\textbf{K}_\texttt{i}\ (\textbf{M}_\texttt{i}\ \varphi))\boldsymbol{\rfloor}$"|\\
%\hspace*{1.5em}\lstinline[language=Isabelle]|by (smt euclidean_def group_S5 pknow_def pimp_def pneg_def pvalid_def|\\
\lstinline[language=Isabelle]|lemma axiom_4: "$\mathcal{A}$ i $\Longrightarrow \boldsymbol{\lfloor}(\textbf{K}_\texttt{i}\ \varphi) \boldsymbol{\rightarrow}(\textbf{K}_\texttt{i}\ (\textbf{K}_\texttt{i}\ \varphi))\boldsymbol{\rfloor}$"|\\
\hspace*{1.5em}\lstinline[language=Isabelle]|by (meson group_S5 transitive_def)|\\
\lstinline[language=Isabelle]|lemma axiom_5: "$\mathcal{A}$ i $\Longrightarrow \boldsymbol{\lfloor}(\boldsymbol{\neg} \textbf{K}_\texttt{i}\ \varphi) \boldsymbol{\rightarrow}(\textbf{K}_\texttt{i}\ \boldsymbol{(\neg} \textbf{K}_\texttt{i}\ \varphi))\boldsymbol{\rfloor}$|\\
\hspace*{1.5em}\lstinline[language=Isabelle]|by (meson euclidean_def group_S5)|\\
\lstinline[language=Isabelle]|$\textcolor{gray}{\text{(*Reduction Axioms*)}}$|\\
\lstinline[language=Isabelle]|lemma atomic_permanence: "$\boldsymbol{\lfloor}(\boldsymbol{[!}\varphi\boldsymbol{]}^\texttt{A}\texttt{p})\boldsymbol{\rightarrow}(\varphi \boldsymbol{\rightarrow} ^\texttt{A}\texttt{p})\boldsymbol{\rfloor}$ by auto|\\
%\hspace*{1.5em}\lstinline[language=Isabelle]|by auto|\\
\lstinline[language=Isabelle]|lemma conjunction: "$\boldsymbol{\lfloor}(\boldsymbol{[!}\varphi\boldsymbol{]}(\psi \boldsymbol{\wedge} \chi)) \boldsymbol{\leftrightarrow} ((\boldsymbol{[!}\varphi\boldsymbol{]}\psi) \boldsymbol{\wedge} (\boldsymbol{[!}\varphi\boldsymbol{]}\chi))\boldsymbol{\rfloor}$ by auto|\\
%\hspace*{1.5em}\lstinline[language=Isabelle]|by auto|\\
\lstinline[language=Isabelle]|lemma partial_functionality: "$\boldsymbol{\lfloor}(\boldsymbol{[!}\varphi\boldsymbol{]\neg}\psi)\boldsymbol{\leftrightarrow}(\varphi \boldsymbol{\rightarrow} (\boldsymbol{\neg [!}\varphi\boldsymbol{]}\psi))\boldsymbol{\rfloor}$ by auto|\\
%\hspace*{1.5em}\lstinline[language=Isabelle]|by auto|\\
\lstinline[language=Isabelle]|lemma action_knowledge: "$\mathcal{A}$ i $\Longrightarrow\boldsymbol{\lfloor}(\boldsymbol{[!}\varphi\boldsymbol{]}(\textbf{K}_\texttt{i}\ \psi)) \boldsymbol{\leftrightarrow} (\varphi \boldsymbol{\rightarrow} (\textbf{K}_\texttt{i}\ (\varphi \boldsymbol{\rightarrow} (\boldsymbol([!\varphi\boldsymbol{]}\psi))))\boldsymbol{\rfloor}$|\\
\hspace*{1.5em}\lstinline[language=Isabelle]|by auto|\\
\lstinline[language=Isabelle]|lemma "$\boldsymbol{\lfloor}(\boldsymbol{[!}\varphi\boldsymbol{]}\textbf{C}\boldsymbol{\llparenthesis}\psi\boldsymbol{|}\chi\boldsymbol{\rrparenthesis}) \boldsymbol{\leftrightarrow} (\varphi \boldsymbol{\rightarrow} \textbf{C}\boldsymbol{\llparenthesis [!}\varphi\boldsymbol{]}\psi\boldsymbol{|[!}\varphi\boldsymbol{]}\chi\boldsymbol{\rrparenthesis})\boldsymbol{\rfloor}$|\\
\hspace*{1.5em}\lstinline[language=Isabelle]|by (smt intersection_rel_def sub_rel_def tc_def transitive_def)|\\
\lstinline[language=Isabelle]|$\textcolor{gray}{\text{(*Axiom schemes for RCK*)}}$|\\
\lstinline[language=Isabelle]|lemma C_normality:  "$\boldsymbol{\lfloor}(\textbf{C}\boldsymbol{\llparenthesis}\chi\boldsymbol{|}\varphi\boldsymbol{\rightarrow}\psi\boldsymbol{\rrparenthesis}) \boldsymbol{\rightarrow} (\textbf{C}\boldsymbol{\llparenthesis}\chi\boldsymbol{|}\varphi\boldsymbol{\rrparenthesis\rightarrow}\textbf{C}\boldsymbol{\llparenthesis}\chi\boldsymbol{|}\psi\boldsymbol{\rrparenthesis})\boldsymbol{\rfloor}$|\\
\hspace*{1.5em}\lstinline[language=Isabelle]|unfolding Defs by blast|\\
\lstinline[language=Isabelle]|lemma mix_axiom1: "$\boldsymbol{\lfloor}\textbf{C}\boldsymbol{\llparenthesis}\chi\boldsymbol{|}\varphi\boldsymbol{\rrparenthesis \rightarrow} (\textbf{E}_\mathcal{A} (\chi \boldsymbol{\rightarrow} (\varphi \boldsymbol{\wedge}\textbf{C}\boldsymbol{\llparenthesis}\chi\boldsymbol{|}\psi\boldsymbol{\rrparenthesis})))\boldsymbol{\rfloor}$|\\
\hspace*{1.5em}\lstinline[language=Isabelle]|unfolding Defs by metis|\\
\lstinline[language=Isabelle]|lemma mix_axiom2: "$\boldsymbol{\lfloor} (\textbf{E}_\mathcal{A}\ (\chi \boldsymbol{\rightarrow} (\varphi \boldsymbol{\wedge} \textbf{C}\boldsymbol{\llparenthesis}\chi\boldsymbol{|}\psi\boldsymbol{\rrparenthesis}))) \boldsymbol{\rightarrow}\textbf{C}\boldsymbol{\llparenthesis}\chi\boldsymbol{|}\varphi\boldsymbol{\rrparenthesis\rfloor}$|\\
\hspace*{1.5em}\lstinline[language=Isabelle]|unfolding Defs sledgehammer $\textcolor{gray}{\text{(*timeout*)}}$|\\
\lstinline[language=Isabelle]|lemma induction_axiom1: "$\boldsymbol{\lfloor} (\textbf{E}_\mathcal{A}\ (\chi \boldsymbol{\rightarrow} \varphi)) \boldsymbol{\wedge} \textbf{C}\boldsymbol{\llparenthesis}\chi\boldsymbol{|}\varphi\boldsymbol{\rightarrow} (\textbf{E}_\mathcal{A}\ (\chi \boldsymbol{\rightarrow} \varphi))\boldsymbol{\rrparenthesis})\boldsymbol{\rightarrow} \textbf{C}\boldsymbol{\llparenthesis}\chi\boldsymbol{|}\varphi\boldsymbol{\rrparenthesis\rfloor}$|\\
\hspace*{1.5em}\lstinline[language=Isabelle]|unfolding Defs sledgehammer $\textcolor{gray}{\text{(*timeout*)}}$|\\
\lstinline[language=Isabelle]|lemma induction_axiom2: "$\boldsymbol{\lfloor}\textbf{C}\boldsymbol{\llparenthesis}\chi\boldsymbol{|}\varphi\boldsymbol{\rrparenthesis \rightarrow} (\textbf{E}_\mathcal{A}\ (\chi \boldsymbol{\rightarrow} \varphi)) \boldsymbol{\wedge}\textbf{C}\boldsymbol{\llparenthesis}\chi\boldsymbol{|}\varphi\boldsymbol{\rightarrow} (\textbf{E}_\mathcal{A}\ (\chi \boldsymbol{\rightarrow} \varphi))\boldsymbol{\rrparenthesis})\boldsymbol{\rfloor}$|\\
\hspace*{1.5em}\lstinline[language=Isabelle]|unfolding Defs by smt|\\
\lstinline[language=Isabelle]|$\textcolor{gray}{\text{(*Rules of Inference*)}}$|\\
\lstinline[language=Isabelle]|lemma announcement_nec: assumes 1: "$\boldsymbol{\lfloor}\varphi\boldsymbol{\rfloor}$" shows "$\boldsymbol{\lfloor [!} \psi\boldsymbol{]}\varphi\boldsymbol{\rfloor}$" using 1 by auto|\\
%\hspace*{1.5em}\lstinline[language=Isabelle]|using 1 by auto|\\
\lstinline[language=Isabelle]|lemma rkc_necessitation: assumes 1: "$\boldsymbol{\lfloor}\varphi\boldsymbol{\rfloor}$" shows "$\boldsymbol{\lfloor C\llparenthesis}\chi\boldsymbol{|}\varphi\boldsymbol{\rrparenthesis\rfloor}$"|\\
\hspace*{1.5em}\lstinline[language=Isabelle]|using 1 by (metis intersection_rel_def sub_rel_def tc_def transitive_def)|\\

\subsection{Exploring Failures of Uniform Substitution}

The following principles are examples of sentences that are \textit{valid} for eternal sentences $p$, but not \textit{schematically valid} \cite{holliday2013}. 

\begin{enumerate}
    \setlength\itemsep{.5em}
	\item \underline{$p \rightarrow \neg [!p](\neg p)$}
	
	\lstinline[language=Isabelle]|lemma "$\boldsymbol{\lfloor}^\texttt{A}\texttt{p} \boldsymbol{\rightarrow} \boldsymbol{\neg [!}^\texttt{A}\texttt{p}\boldsymbol{]}(\boldsymbol{\neg}^\texttt{A}\texttt{p}) \boldsymbol{\rfloor}$ by simp|\\
	%\hspace*{1.5em}\lstinline[language=Isabelle]|by simp|\\
	\lstinline[language=Isabelle]|lemma "$\boldsymbol{\lfloor}\varphi \boldsymbol{\rightarrow} \boldsymbol{\neg [!}\varphi\boldsymbol{]}(\boldsymbol{\neg}\varphi) \boldsymbol{\rfloor}$ nitpick oops $\textcolor{gray}{\text{(*countermodel found*)}}$|
	%$\textcolor{gray}{\text{(*countermodel found*)}}$|
	%\hspace*{1.5em}\lstinline[language=Isabelle]|nitpick oops $\textcolor{gray}{\text{(*countermodel found*)}}$|
	
	\item \underline{$p \rightarrow \neg [!p](\neg K_i p)$}
	
	\lstinline[language=Isabelle]|lemma "$\boldsymbol{\lfloor}^\texttt{A}\texttt{p} \boldsymbol{\rightarrow} \boldsymbol{\neg [!}^\texttt{A}\texttt{p}\boldsymbol{]}(\boldsymbol{\neg}\textbf{K}_\texttt{a}\, ^\texttt{A}\texttt{p}) \boldsymbol{\rfloor}$ by simp|\\
	%\hspace*{1.5em}\lstinline[language=Isabelle]|by simp|\\
	\lstinline[language=Isabelle]|lemma "$\boldsymbol{\lfloor}\varphi \boldsymbol{\rightarrow} \boldsymbol{\neg [!}\varphi\boldsymbol{]}(\boldsymbol{\neg}\textbf{K}_\texttt{a}\, \varphi) \boldsymbol{\rfloor}$ nitpick oops $\textcolor{gray}{\text{(*countermodel found*)}}$| %$\textcolor{gray}{\text{(*countermodel found*)}}$|
	%\hspace*{1.5em}\lstinline[language=Isabelle]|nitpick oops $\textcolor{gray}{\text{(*countermodel found*)}}$|
	
	\item \underline{$p \rightarrow \neg [!p](p \wedge \neg K_i p)$}
	
	\lstinline[language=Isabelle]|lemma "$\boldsymbol{\lfloor}^\texttt{A}\texttt{p} \boldsymbol{\rightarrow} \boldsymbol{\neg [!}^\texttt{A}\texttt{p}\boldsymbol{]}(^\texttt{A}\texttt{p} \boldsymbol{\wedge \neg} \textbf{K}_\texttt{a}\, ^\texttt{A}\texttt{p}) \boldsymbol{\rfloor}$ by simp|\\
	%\hspace*{1.5em}\lstinline[language=Isabelle]|by simp|\\
	\lstinline[language=Isabelle]|lemma "$\boldsymbol{\lfloor}\varphi \boldsymbol{\rightarrow} \boldsymbol{\neg [!}\varphi\boldsymbol{]}(\varphi \boldsymbol{\wedge}\boldsymbol{\neg} \textbf{K}_\texttt{a}\, \varphi) \boldsymbol{\rfloor}$ nitpick oops $\textcolor{gray}{\text{(*countermodel found*)}}$|
	%$\textcolor{gray}{\text{(*countermodel found*)}}$|
	%\hspace*{1.5em}\lstinline[language=Isabelle]|nitpick oops $\textcolor{gray}{\text{(*countermodel found*)}}$|
	
	\item \underline{$(p \wedge \neg K_i p) \rightarrow \neg [!p \wedge \neg K_i p] (p \wedge \neg K_i p)$}
	
	\lstinline[language=Isabelle]|lemma "$\boldsymbol{\lfloor}(^\texttt{A}\texttt{p} \boldsymbol{\wedge} \boldsymbol{\neg}\textbf{K}_\texttt{a}\, ^\texttt{A}\texttt{p}) \boldsymbol{\rightarrow} \boldsymbol{\neg}\boldsymbol{[!}^\texttt{A}\texttt{p} \boldsymbol{\wedge} \boldsymbol{\neg}\textbf{K}_\texttt{a}\, ^\texttt{A}\texttt{p}\boldsymbol{]}(^\texttt{A}\texttt{p} \boldsymbol{\wedge} \boldsymbol{\neg}\textbf{K}_\texttt{a}\, ^\texttt{A}\texttt{p})\boldsymbol{\rfloor}$ by blast|\\
	%\hspace*{1.5em}\lstinline[language=Isabelle]|by blast|\\
	\lstinline[language=Isabelle]|lemma "$\boldsymbol{\lfloor}(\varphi \boldsymbol{\wedge} \boldsymbol{\neg}\textbf{K}_\texttt{a}\, \varphi) \boldsymbol{\rightarrow} \boldsymbol{\neg}\boldsymbol{[!}\varphi \boldsymbol{\wedge} \boldsymbol{\neg}\textbf{K}_\texttt{a}\, \varphi\boldsymbol{]}(\varphi \boldsymbol{\wedge} \boldsymbol{\neg}\textbf{K}_\texttt{a}\, \varphi)\boldsymbol{\rfloor}$ nitpick oops $\textcolor{gray}{\text{(*ctm. fd.*)}}$|
	%\hspace*{1.5em}\lstinline[language=Isabelle]|nitpick oops $\textcolor{gray}{\text{(*countermodel found*)}}$|
	
	\item \underline{$K_i p \rightarrow \neg [!p](\neg K_i p)$}

	\lstinline[language=Isabelle]|lemma "$\boldsymbol{\lfloor}(\textbf{K}_\texttt{a}\, ^\texttt{A}\texttt{p}) \boldsymbol{\rightarrow} \boldsymbol{\neg [!}^\texttt{A}\texttt{p}\boldsymbol{]}(\boldsymbol{\neg}\textbf{K}_\texttt{a}\, ^\texttt{A}\texttt{p}) \boldsymbol{\rfloor}$ using group_S5 reflexive_def by auto|\\
	%\hspace*{1.5em}\lstinline[language=Isabelle]|using group_S5 reflexive_def by auto|\\
	\lstinline[language=Isabelle]|lemma "$\boldsymbol{\lfloor}(\textbf{K}_\texttt{a} \varphi) \boldsymbol{\rightarrow} \boldsymbol{\neg [!}\varphi\boldsymbol{]}(\boldsymbol{\neg}\textbf{K}_\texttt{a}\, \varphi) \boldsymbol{\rfloor}$ nitpick oops $\textcolor{gray}{\text{(*countermodel found*)}}$|
	%\hspace*{1.5em}\lstinline[language=Isabelle]|nitpick oops $\textcolor{gray}{\text{(*countermodel found*)}}$|
	
	\item \underline{$K_i p \rightarrow \neg [!p](p \wedge \neg K_i p)$}

	\lstinline[language=Isabelle]|lemma "$\boldsymbol{\lfloor}(\textbf{K}_\texttt{a}\, ^\texttt{A}\texttt{p}) \boldsymbol{\rightarrow} \boldsymbol{\neg [!}^\texttt{A}\texttt{p}\boldsymbol{]}(^\texttt{A}\texttt{p} \boldsymbol{\wedge} \boldsymbol{\neg} \textbf{K}_\texttt{a}\, ^\texttt{A}\texttt{p}) \boldsymbol{\rfloor}$ using group_S5 reflexive_def by auto|\\
	%\hspace*{1.5em}\lstinline[language=Isabelle]|using group_S5 reflexive_def by auto|\\
	\lstinline[language=Isabelle]|lemma "$\boldsymbol{\lfloor}(\textbf{K}_\texttt{a}\, \varphi) \boldsymbol{\rightarrow} \boldsymbol{\neg [!}\varphi\boldsymbol{]}(\varphi \boldsymbol{\wedge} \boldsymbol{\neg} \textbf{K}_\texttt{a}\, \varphi) \boldsymbol{\rfloor}$ nitpick oops $\textcolor{gray}{\text{(*countermodel found*)}}$|\\
	%\hspace*{1.5em}\lstinline[language=Isabelle]|nitpick oops $\textcolor{gray}{\text{(*countermodel found*)}}$|
\end{enumerate}

%\hl{As expected, whenever Nitpick got invoked it correctly reported counterexamples.}

\subsection{Example Application: The Wise Men Puzzle}

The Wise Men puzzle is a interesting riddle in epistemic reasoning.
It is well suited to demonstrate epistemic actions in a multi-agent scenario.
%Baldoni gave a formulation for this in his PhD thesis \cite{baldoni1998normal}.
%This formulation later got embedded into Isabelle/HOL by Benzmüller \cite{J41}.
Baldoni~\cite{baldoni1998normal} gave a formulation for this, which later got embedded into Isabelle/HOL by Benzmüller~\cite{J41,J44}.
In the following implementation these results will be used as a stepping stone.

First the riddle is recited, and then we go into detail on how the uncertainties of all three agents change.
The reader is invited to try to solve the riddle on her own before continuing with the analysis. 

\begin{quote}\it \small
	Once upon a time, a king wanted to find the wisest out of his three wisest men.
	He arranged them in a circle and told them that he would put a white or a black spot on their foreheads and that one of the three spots would certainly be white.
	The three wise men could see and hear each other but, of course, they could not see their faces reflected anywhere.
	The king, then, asked each of them [sequentially] to find out the color of his own spot.
	After a while, the wisest correctly answered that his spot was white.
\end{quote}

\noindent
The already existing encoding by Benzm\"uller puts a particular emphasis on the adequate modeling of common knowledge.
Here, this solution will be enhanced by the public announcement operator.
Consequently, common knowledge will not be statically stated after each iteration, but a dynamic approach is used for this.

Before we can evaluate the knowledge of the first wise man we need to formulate the initial circumstances and background knowledge.
Let \texttt{a}, \texttt{b} and \texttt{c} be the wise men.
It is common knowledge, that each wise man can see the foreheads of the other wise men.
The only doubt a wise man has, is whether he has a white spot on his own forehead or not.
%This is common knowledge among the group.
Additionally, it is common knowledge that at least one of the three wise men has a white spot on his forehead.
The rules of the riddle are embedded as follows:\footnote{One might also add axioms of the form $\lfloor \texttt{\textbf{C}}_\mathcal{A} \ (^\text{A}\texttt{ws x}) \boldsymbol{\rightarrow} \texttt{\textbf{K}}_\texttt{y} (^\text{A}\texttt{ws x}) ) \rfloor"$ for \texttt{x}, \texttt{y} $\in \mathcal{A}$. This is not necessary as we will see in the proof found using Isabelle/HOL.}

%\begin{lstlisting}[frame=None]
%consts ws :: "$\alpha\Rightarrow\sigma$"
%axiomatization where
%$\textcolor{gray}{\text{(* Common knowledge: at least one of a, b and c has a white spot *)}}$
%WM1: "$\lfloor \textbf{C}_\mathcal{A} \ (^\text{A}\text{ws a} \ \boldsymbol{\vee} \ ^\text{A}\text{ws b} \ \boldsymbol{\vee} \ ^\text{A}\text{ws c})$"
%$\textcolor{gray}{\text{(* Common knowledge: if x has a not white spot then y know this *)}}$
%WM2: "$\mathcal{A}\ \text{x} \ \boldsymbol{\wedge} \ \mathcal{A}\ \text{y} \ \boldsymbol{\wedge}$ x $\not =$ y $\Longrightarrow \lfloor \textbf{C}_\mathcal{A} \ (\boldsymbol{\neg} (^\text{A}\text{ws x}) \boldsymbol{\rightarrow} \textbf{K}_\text{y} (\boldsymbol{\neg} (^\text{A}\text{ws x}) )) \rfloor$"
%\end{lstlisting}

%\begin{lstlisting}[frame=single]
\begin{lstlisting}[frame=None]
consts ws :: "$\alpha\Rightarrow\sigma$"
axiomatization where
  $\textcolor{gray}{\text{(* Common knowledge: at least one of a, b and c has a white spot *)}}$
  WM1: "$\lfloor \textbf{C}_\mathcal{A} \ (^\text{A}\text{ws a} \ \boldsymbol{\vee} \ ^\text{A}\text{ws b} \ \boldsymbol{\vee} \ ^\text{A}\text{ws c})\textcolor{red}{\rfloor}$"
  $\textcolor{gray}{\text{(* Common knowledge: if x has not a white spot then y know this *)}}$
  WM2ab: "$\lfloor \textbf{C}_\mathcal{A} \ (\boldsymbol{\neg} (^\text{A}\text{ws a}) \boldsymbol{\rightarrow} \textbf{K}_\text{b} (\boldsymbol{\neg} (^\text{A}\text{ws a}) )) \rfloor$"
  WM2ac: "$\lfloor \textbf{C}_\mathcal{A} \ (\boldsymbol{\neg} (^\text{A}\text{ws a}) \boldsymbol{\rightarrow} \textbf{K}_\text{c} (\boldsymbol{\neg} (^\text{A}\text{ws a}) )) \rfloor$"
  WM2ba: "$\lfloor \textbf{C}_\mathcal{A} \ (\boldsymbol{\neg} (^\text{A}\text{ws b}) \boldsymbol{\rightarrow} \textbf{K}_\text{a} (\boldsymbol{\neg} (^\text{A}\text{ws b}) )) \rfloor$"
  WM2bc: "$\lfloor \textbf{C}_\mathcal{A} \ (\boldsymbol{\neg} (^\text{A}\text{ws b}) \boldsymbol{\rightarrow} \textbf{K}_\text{c} (\boldsymbol{\neg} (^\text{A}\text{ws b}) )) \rfloor$"
  WM2ca: "$\lfloor \textbf{C}_\mathcal{A} \ (\boldsymbol{\neg} (^\text{A}\text{ws c}) \boldsymbol{\rightarrow} \textbf{K}_\text{a} (\boldsymbol{\neg} (^\text{A}\text{ws c}) )) \rfloor$"
  WM2cb: "$\lfloor \textbf{C}_\mathcal{A} \ (\boldsymbol{\neg} (^\text{A}\text{ws c}) \boldsymbol{\rightarrow} \textbf{K}_\text{b} (\boldsymbol{\neg} (^\text{A}\text{ws c}) )) \rfloor$"
\end{lstlisting}

\noindent
Now the king asks $a$ whether he knows if he has a white spot or not.
Assume that $a$ publicly answers that he does not. This is a public announcement of the form: $\neg (\texttt{K}_\texttt{a} (^\texttt{A}\texttt{ws a})) \vee \texttt{K}_\texttt{a} \neg (^\texttt{A}\texttt{ws a}))$.
Again, a wise man gets asked by the king whether he knows if he has a white spot or not.
Now its $b$'s turn and assume that $b$ also announces that he does not know whether he has a white spot on his forehead.\footnote{The case where neither $a$ nor $b$ can correctly infer the color of their forehead when being asked by the king is the most challenging case; we only discuss this one here.} 

When asked, $c$ is able to give the right answer, namely that he has a white spot on his forehead.
%This demonstrates, that \textit{statements about ignorance can lead to knowledge}.
We can prove this automatically in Isabelle/HOL:

%\lstinline[language=isabelle]|declare [[smt_solver=cvc4, smt_oracle]]|
%\lstinline[language=isabelle]|theorem whitespot_c:|\\
%\hspace*{1.5em}\lstinline[language=isabelle]|"$\lfloor \boldsymbol{[!}\boldsymbol{\neg} ((\textbf{K}_\text{a} (^\text{A}\text{ws a})) \boldsymbol{\vee} (\textbf{K}_\text{a} (\boldsymbol{\neg}(^\text{A}\text{ws a})))\boldsymbol{]}($|\\
%\hspace*{3.5em}\lstinline[language=isabelle]|$\boldsymbol{[!}\boldsymbol{\neg} ((\textbf{K}_\text{b} (^\text{A}\text{ws b})) \boldsymbol{\vee} (\textbf{K}_\text{b} (\boldsymbol{\neg}(^\text{A}\text{ws b})))\boldsymbol{]}($|\\
%\hspace*{5em}\lstinline[language=isabelle]|$\textbf{K}_\text{c} (^\text{A}\text{ws c})))\rfloor$"|\\
%\hspace*{1.5em}\lstinline[language=isabelle]|using WM2ba WM2ca WM2cb group_S5|\\
%\hspace*{1.5em}\lstinline[language=isabelle]|unfolding reflexive_def intersection_rel_def union_rel_def sub_rel_def|\\
%\hspace*{2.5em}\lstinline[language=isabelle]| tc_def EVR_def by smt|

% WM2ac war nicht notwendig
\begin{lstlisting}[frame=None]
theorem whitespot_c:
 "$\lfloor \boldsymbol{[!}\boldsymbol{\neg} ((\textbf{K}_\text{a} (^\text{A}\text{ws a})) \boldsymbol{\vee} (\textbf{K}_\text{a} (\boldsymbol{\neg}(^\text{A}\text{ws a})))\boldsymbol{]}(\boldsymbol{[!}\boldsymbol{\neg} ((\textbf{K}_\text{b} (^\text{A}\text{ws b})) \boldsymbol{\vee} (\textbf{K}_\text{b} (\boldsymbol{\neg}(^\text{A}\text{ws b})))\boldsymbol{]}(\textbf{K}_\text{c} (^\text{A}\text{ws c})))\rfloor$"
  using WM1 WM2ba WM2ca WM2cb group_S5
  unfolding reflexive_def intersection_rel_def
    union_rel_def sub_rel_def tc_def
  by smt
\end{lstlisting}

\section{Comparison with Related Work}
In related work \cite{BenthemEGS18}, van Benthem, van Eijck and colleagues have studied a  \textit{``faithful representation of DEL [dynamic epistemic logic] models as so-called knowledge structures that allow for symbolic model checking''.}  The authors show that such an approach enables efficient and effective reasoning in epistemic scenarios with state-of-the-art \hl{Binary Decision Diagram (BDD)} reasoning technology, outperforming other existing methods \cite{DEMO,DEMO-S5} to automate DEL reasoning.
%\hl{It is also possible to re-formalize dynamic epistemic terms in temporal epistemic terms [cite] and then use the model checkers MCK [cite] or MCMAS [cite].}
Further related work 
\hl{\cite{van2006model} demonstrates how dynamic epistemic terms can be formalized in temporal epistemic terms to apply the model checkers MCK \cite{gammie2004mck} or MCMAS \cite{raimondi2004verification}.}
Our approach differs in various respects, incuding: 

\begin{description}
\item[External vs.~internal representation transformation:] Instead of writing external (e.g~Haskell-)code to realize the required conversions from DEL into Boolean representations, we work with logic-internal conversions into HOL, provided in form of a set of equations stated in HOL itself (thereby heavily exploiting the virtues of $\lambda$-abstraction and $\lambda$-conversion). Our encoding is concise (only about 50 lines in Isabelle/HOL) and human readable.

\item[Meta-logical reasoning:]
    Since our conversion ``code'' is provided within the (meta-)logic environment itself, the conversion becomes better controllable and even amenable to formal verification. Moreover, as we have also demonstrated in this paper, meta-logical studies about the embedded logics and their embedding in HOL are well-supported in our approach.

\item[Scalability beyond propositional reasoning:]
    Real world applications often require differentiation between entities/individuals, their properties and functions defined on them, and quantification over entities, or even properties and functions, supports generic statements that are not supported in propositional DEL.  The shallow semantical embedding approach, in contrast, very naturally scales for first-order and  higher-order extensions of the embedded logics; for more details on this we refer to \cite{J41,J44} and the references therein.

\item[Reuse of automated theorem proving and model finding technology:] Both approaches reuse state-of-the-art automated reasoning technology. In our case this includes world-leading first-order and higher-order theorem provers and model finders already integrated with Isabelle/HOL \cite{sledgehammer}. These tools in turn internally collaborate with latest SMT and SAT solving technology. The burden to organize and orchestrate the technical communication with and between these tools is taken away from us by reuse of respective solutions as already provided in Isabelle/HOL (and recursively also within the integrated theorem provers). Well established and robustly supported language formats (e.g.~TPTP syntax, \url{http://www.tptp.org}) are reused in these nested transformations. These cascades of already supported logic transformations 
%the availability of these different, tool-internal transformations (into HOL, first-order and eventually Boolean representations) 
are one reason why our embedding approach readily scales for automating reasoning beyond just propositional DEL. 
\end{description}

We are convinced, as evidenced by the above discussion, that our approach is particularly well suited for the exploration and rapid prototyping of new logics (and logic combinations) and their embeddings in HOL, and for the study of their meta-logical properties, in particular, when it comes to first-order and higher-order extensions of DEL. 
At the same time we share with the related work by van Benthem, van Eijck a and colleagues  a deep interest in practical (object-level) applications, and therefore practical reasoning performance is obviously also of high relevance. In this regard, however, we naturally assume a performance loss in comparison to hand-crafted, specialist solutions. Previous studies in the context of first-order modal logic theorem proving nevertheless have shown that this is not always the case \cite{C62}. Future work therefore includes the conduction of comparative performance studies in which the work presented in this paper is compared with the existing alternative approaches. % In the long run, we are convinced, both approaches could eventually be combined within a proof assistant such as Isabelle/HOL.

\section{Conclusion}
A shallow semantical embedding of public announcement logic with relativized common knowledge in classical higher-order logic has been presented, and our implementation of this embedding in Isabelle/HOL delivers results as expected. In particular, we have shown how model-changing behaviour can be adequately and elegantly addressed in our embedding approach.
With reference to uniform substitution, we saw that our embedding
enables the study of meta-logical properties of public  announcement logic, and object-level reasoning has been demonstrated by a first time automation of the wise men puzzle encoded in public announcement logic with a relativized common knowledge operator.

Further work includes the provision of proofs for the faithfulness of the presented embedding; this should be analogous to prior work, see e.g.~\cite{J31}.

\subsubsection{\hl{Acknowledgments}}

\hl{We thank David Streit, David Fuenmayor and the anonymous reviewers for useful comments, suggestions and feedback to this work.}

\bibliographystyle{abbrv}
\bibliography{chris,bibliography}

\end{document}